\documentclass[journal]{IEEEtran}
\ifCLASSINFOpdf
\else
\fi
\usepackage[utf8x]{inputenc} 
\usepackage{xspace}
\usepackage{amsmath,amssymb,amsfonts}
\usepackage{algorithmic}
\usepackage{graphicx}
\usepackage[table,xcdraw]{xcolor}
\usepackage{tabularx}
\usepackage{multirow}
\usepackage[switch]{lineno}
\usepackage{textcomp}
\usepackage{xspace}
\usepackage{blindtext}

\begin{document}
%
\title{A generic self-supervised learning (SSL) framework for representation learning from spectra-spatial feature of unlabeled remote sensing imagery}
%
%
%

\author{Xin~Zhang,
        Liangxiu~Han*
\thanks{Xin Zhang and Liangxiu Han are with the Department of Computing, and Mathematics, Manchester Metropolitan University, Manchester M15GD, UK e-mail: l.han@mmu.ac.uk;.}
\thanks{Manuscript received April 19, 2005; revised September 17, 2014.}}

%
%

\markboth{Journal of \LaTeX\ Class Files,~Vol.~13, No.~9, September~2014}%
{Shell \MakeLowercase{\textit{et al.}}: Bare Demo of IEEEtran.cls for Journals}
%



\maketitle

\begin{abstract}
Remote sensing data has been widely used for various Earth Observation (EO) missions such as land use and cover classification, weather forecasting, agricultural management, and environmental monitoring. Most existing remote sensing data-based models are based on supervised learning that requires large and representative human-labelled data for model training, which is costly and time-consuming. Recently, self-supervised learning (SSL) enables the models to learn a representation from orders of magnitude more unlabelled data. This representation has been proven to boost the performance of downstream tasks and has potential for remote sensing applications. The success of SSL is heavily dependent on a pre-designed pretext task, which introduces an inductive bias into the model from a large amount of unlabelled data. Since remote sensing imagery has rich spectral information beyond the standard RGB colour space, the pretext tasks established in computer vision based on RGB images may not be straightforward to be extended to the multi/hyperspectral domain. To address this challenge, this work has designed a novel SSL framework that is capable of learning representation from both spectra-spatial information of unlabelled data. The framework contains two novel pretext tasks for object-based and pixel-based remote sensing data analysis methods, respectively. Through two typical downstream tasks evaluation (a multi-label land cover classification task on Sentienl-2 multispectral datasets and a ground soil parameter retrieval task on hyperspectral datasets), the results demonstrate that the representation obtained through the proposed SSL achieved a significant improvement in model performance. By comparing with the currently available SSL methods, the proposed method that emphasizes both spectral and spatial features outperform these existing SSL methods on multi and hyperspectral remote sensing datasets.
\end{abstract}


%
\IEEEpeerreviewmaketitle

\section{Introduction}
%
%
%
%
\IEEEPARstart{E}{arth} observation through remote sensing data provides an unbiased, uninterrupted, and borderless view of human activities and natural processes. Through exploiting data collected from various aircraft and satellite systems equipped with multi/hyper spectral sensors ranging from medium and very high spatial resolution, together with advanced data analysis/ machine learning,  people can gain digital information and insight to guide multiple applications concerning any corner of the planet \cite{ban2015global}. In detail, the remote sensing data analysis is essential for many applications such as environmental monitoring, natural resource management, disaster response, urban planning, climate change studies\cite{li_recent_2020,osco_review_2021}. With the rapid development of sensor technology, the complexity of remote sensing data has increased significantly due to the rapid improvement in its spatial and spectral resolution, which poses challenge to the remote sensing data analysis\cite{ghamisi_advanced_2017}. 

In general, the existing remote sensing data analysis methods usually contain two fundamental components: data processing and feature extraction. Depending on the spatial and spectral resolution of data, the data processing methods can be broadly divided into pixel-based and object-based methods. Specifically, the most commonly used pixel-based methods take each individual pixel as input and utilize the rich spectral information for subsequent feature extraction tasks \cite{richards_remote_2006}. It is suitable for low to medium spatial resolution remote sensing data. As the spatial resolution of the data increases, individual pixels are no longer able to cover an object target on the ground. The object-based methods is introduced to segment an image into objects containing spectral and spatial information (e.g. shape/geometry and structure) \cite{chen_geographic_2018} for subsequent feature extraction and analysis. 

Regarding feature extraction, the traditional machine learning methods, such as Classification and Regression Tree (CART) \cite{pal_assessment_2003}; Support Vector Machine (SVM) \cite{cortes_support-vector_1995}, and Random Forest (RF) \cite{breiman_random_2001} have widely been used for extracting features from remote sensed data for various tasks, such as land cover classification \cite{pal_random_2005}, carbon emission, biomass estimation \cite{safari_comparative_2017,singh_remote_2022}. Recent development of deep learning methods \cite{garcia-garcia_review_2017,zhang_how_2020}, such as convolutional neural networks, has shown promising in remote sensing applications and have achieved state-of-the-art performance \cite{ball_comprehensive_2017} since the convolution operation is able to capture spatial-spectral information \cite{romero_unsupervised_2016}. 

However, most of existing machine/deep learning-based methods are supervised learning, which requires extensive annotated  datasets. Acquiring such well-labelled data is labour intensive and time consuming. Recently, self-supervised learning has been proposed to learn patterns from unlabelled data and has been effectively applied in many fields such as computer vision\cite{hatano_image_2020,li_multi-task_2020}, natural language processing \cite{lan_albert_2019,leiter_chatgpt_2023}, and object detection \cite{misra_self-supervised_2020,mitash_self-supervised_2017}. Essentially, the self-supervised learning consists of two steps: firstly, it has an auxiliary or pretext task using pseudo-labels (i.e., auto-generated labels) to help initialize the model parameters, which are then used for boosting downstream tasks such as classification, segmentation and object detection. 

Until now, a few SSLs have been applied directly on remote sensing applications\cite{alosaimi_self-supervised_2023,tao_remote_2022} including land use/cover mapping \cite{zhao_when_2020}, change detection \cite{dong_self-supervised_2020} and Nitrogen prediction \cite{zhang_self-supervised_2022}. Most of the existing self-supervised for remote sensing analysis are straightforward extension of methods in computer vision domain. These pretext tasks are designed to learn the spatial features from RGB data, such as inpainting of the data \cite{he_masked_2021}, disrupt the spatial order of the data and random rotate the image \cite{komodakis_unsupervised_2018}. However, given remote sensing imagery contains spectral bands beyond the standard RGB colour space (i.e., including both spectral and spatial information), it is insufficient to directly extend the pretext task learning from RGB images to remote sensing data. To the best of our knowledge, there is currently no pretext task designed for spectral-spatial information extraction. Therefore, this work proposes a generic SSL framework for both spatial and spectral feature learning from label-free remote sensing data. This SSL could directly learn a high-level representation from the remote sensing image to promote the performance of both pixel and object-based downstream tasks. The main contributions of this work are as follows:
\begin{enumerate}
  \item We propose a generic SSL framework for both pixel-based and object-based remote sensing applications. Two novel pretext tasks are proposed. One is used to reconstruct the spectral profile from the masked data, which can be used to extract a representation of pixel information and improve the downstream task performance belonging to pixel-based analysis. The other pretext task is proposed to identify objects from multiple views of the same object. These multiple views including global views, local views, and spectral views, are derived from extensive spatial and spectral transformations of the data to allow the model to learn representations from the spatial-spectral information of the data. These representation can be used to improve the downstream task performance belonging to object-based analysis.
  \item We demonstrate that the proposed SSL is a novel way to learn representation from unlabelled large-scale remote sensing data. This proposed SSL method is applied to two downstream tasks on large multispectral and hyperspectral remote sensing datasets. One is a multi-label land cover classification on Sentienl-2 multispectral datasets and the other is a ground soil parameter retrieval on hyperspectral datasets. We also compared the proposed methods with the existing SSL frameworks. The results show the proposed SSL method emphasizes the spectral and spatial features in remote sensing data with higher performance than the other three methods.
  \item We analyse the impact of spatial-spectral features on the proposed SSL performance and visualize the features learned by SSL, which contribute to a deeper understanding of what would make a self-supervised feature representation useful for remote sensing data analysis.

\end{enumerate}

\section{Related work}

\subsection{Remote sensing analysis methods}

In recent years, the amount of available remote sensing data has increased significantly. The spatial and spectral resolution of the remote sensing data has also increased. It brings challenges to remote sensing analysis methods. Unlike the conventional digital imagery, which captures electromagnetic emissions with only three bands (Red, Green, and Blue) in the visible spectrum, remote sensing imagery has a cube form often with multiple bands \cite{imani_overview_2020}, covering a wider range of spectra, including the visible spectrum, infrared spectrum, and radio wave range. In most remote sensing imagery analysis methods, the spectral information of each image pixel, made up of hundreds of spectral bands, acts as an important role. Another fundamental feature of the remote sensing data is spatial information, which normally includes such as the texture, shape, and edges of the ground object. In most remote sennsing analysis methods, how to extract valid features from spectral and spatial information is the most vital component. Generally, the feature extraction methods can be broadly divided into two categories: supervised and self-supervised or unsupervised learning\cite{alosaimi_self-supervised_2023}. 

Supervised learning is the most frequently used feature extraction method from labeled data. Numerous traditional machine learning methods such as the SVM \cite{fauvel_spectral_2007,lee_svm-based_2015}, RF\cite{belgiu_random_2016,pal_random_2005}, and boosted DTs \cite{chasmer_decision-tree_2014,friedl_decision_1997} have been widely used  for the feature extraction of remote sensing data. In recent years, deep learning has shown increasing success in a variety of computational vision tasks and is being used in remote sensing applications \cite{ball_comprehensive_2017,ball_special_2018}. In \cite{zhang_how_2020}, deep learning methods are used in both pixel-based and object-based remote sensing applications and have demonstrated superior performance over traditional machine learning methods. Meanwhile, the authors evalute the performance of a variety of deep learning models in a land cover and object detection tasks. In \cite{brown_dynamic_2022}, Google trained a deep learning model for land cover mapping by using Sentinel-2 10m dataset.This model enables real-time land cover prediction on a global scale. 

However, supervised learning of remote sensing data requires large labelled data for model training. This poses several challenges. One of the big challenges is that manual annotation of big remote sensing data is expensive, time consuming, labour intensive and subject to individual bias. Another major challenge for supervised learning in remote sensing is the location sensitivity of annotations. The accuracy of supervised learning methods relies on the location and distribution of the selected annotation areas, which makes these methods lack transferability.  

The self-supervised learning provides a paradigm to address those challenges that train models with unlabeled data \cite{wang_self-supervised_2022}. In general, the traditional self-supervised method used in remote sensing applications normally refers to cluster pixels in a dataset based on statistics only, without any user-defined training classes \cite{bruzzone2001unsupervised,congalton1991review}. The two most frequently used algorithms are ISODATA \cite{ball_isodata:_1965}and K-Means \cite{kanungo_efficient_2002}.  However, these traditional SSL methods are designed for clustering, grouping and dimensional reduction \cite{zhang_crop_2016, zhang_spectralspatial_2016}, which do not extract feature for futher analysis. In recent years, a new SSL research trend has emerged to learn representation without labels with deep learning models. These representations can be used to boost the performance of downstream applications, which has great potential for remote sensing applications.

\subsection{Self-supervised learning on remote sensing (SSL)}

In general, Self-supervised learning (SSL) involves two tasks: a self-supervised pretext task, and real downstream tasks. The pretext task aims to train a network by optimizing this objective in a self supervised manner using pseudo-labels (i.e. auto-generated labels) of unlabeled data to help initialize the model parameters. Through carefully designed pretext tasks, the network gained the ability to capture high-level representations of the input. Afterwards, the network can be further transferred to supervised downstream tasks for real-world applications.

The success of SSL is heavily dependent on how well a pretext task is designed. The pretext task implicitly introduces an inductive bias into the model learning from a large amount of unlabeled data. If not designed properly, the learning model will only be able to find low-level features, which will be difficult to use for real downstream tasks. Several pretext tasks have been proposed for self-supervised representation learning using visual common sense, such as predicting rotation angle \cite{komodakis_unsupervised_2018}, relative patch position \cite{doersch_unsupervised_2015}, solving jigsaw puzzle games \cite{noroozi_unsupervised_2016}. 

There are two common strategies for pretext design to  achieve different objectives: 1) Generative-based pretext task that reconstructs the input data (Such as discriminating images created from distortion \cite{alexey_discriminative_2016}) $f(x) \rightarrow x$, or predicts a label $c$ that is self-generated from context and data augmentation $f(x) \rightarrow c$,  and  2) Contrastive learning \cite{arora_theoretical_2019} based pretext task that contrasts inputs $x_1$ and $x_2$ that have similar meanings (for example, the encoded features of two different views of the same image should match \cite{caron_emerging_2021,grill_bootstrap_2020}) $\left|f\left(x_1\right)-f\left(x_2\right)\right| \rightarrow 0$. {Table.~\ref{table:1}} summarises the representative approaches for different types of pretext tasks.

\begin{table*}[h]
\caption{A representative collection of pretext tasks in the existing SSL methods.}\label{table:1}
\centering
\resizebox{0.9\textwidth}{!}{%
\begin{tabular}{cll}
\hline
\textbf{CATEGORY}                                    & \textbf{Name}                                           & \textbf{Pretext task}                                                                                \\ \hline
\multirow{7}{*}{\textbf{Generative based}}           & Denoising AE \cite{vincent_stacked_2010}                    & Reconstruct clear image from noisy input                                                             \\
                                                     & Masked AE (MAE) \cite{he_masked_2021}                       & Reconstruct randomly masked patches                                                                  \\
                                                     & GANs \cite{goodfellow_generative_2014}                           & Adversarial training with a generator and a discriminator                                            \\
                                                     & Wasserstein GAN \cite{arjovsky_wasserstein_2017}                 & Train the generator to produce samples that are as close as possible   to the real data distribution \\
                                                     & Relative position \cite{doersch_unsupervised_2015}                 & Predict the relative positions of random patch pairs                                                 \\
                                                     & Rotation \cite{komodakis_unsupervised_2018}                    & Predict the rotation angle of the random rotated image                                               \\
                                                     & puzzle \cite{noroozi_unsupervised_2016}                         & Predict the correct order of the puzzle                                                              \\ \hline
\multirow{4}{*}{\textbf{Contrastive learning based}} & MoCo V1-V3 \cite{chen_improved_2020,chen_empirical_2021,he_momentum_2020} & Store negative samples in a queue and perform momentum updates to the   key code.                    \\
                                                     & SwAV \cite{caron_emerging_2021}                        & Contrastive learning for online clustering                                                           \\
                                                     & BYOL \cite{grill_bootstrap_2020}                                & Average a teacher network with a predictor on top of a teacher encoder                               \\
                                                     & SimSiam \cite{chen_exploring_2020}                            & Explore the simplest contrasting SSL designs                                                         \\ \hline
\end{tabular}
}
\end{table*}

Naturally, these pretext tasks also have been used for remote sensing applications in a self-supervised manner. In \cite{zhao_when_2020}, and \cite{wen_rotation_2021}, the random rotation pretext task is used to learn the representation from RGB and SAR remote sensing data as a generative-based SSL. These representations are finally used to boost remote sensing classification tasks. In \cite{singh_self-supervised_2018} and \cite{tao_remote_2022}, inpainting and relative position pretext tasks are used for segmentation and classification. In recent years, the contrastive learning based SSL has also been widely used in remote sensing \cite{geng_multi-view_2022,rao_transferable_2022,zhang_self-supervised_2022,zhang_semisupervised_2022}. Tile2Vec is the first self-supervised work using contrastive learning for remote sensing image representation learning \cite{jean_tile2vec_2018}. A triple loss is proposed to encourage neighboring patches in one image that are closer and moving tiles that are further away in the spatial space. \cite{hou_hyperspectral_2021} uses SimCLR like contrastive learning to pre-train HSI classification models to reduce the requirement for massive annotations.

It is worth mentioning that the most current pretext tasks are designed for RGB image where the spatial features are the primary features considered. Only a few simple spectral augmentations \cite{duan_self-supervised_2022,zhu_sc-eadnet_2022} are used for view generation. Remote sensing imagery contains rich spectral bands beyond the standard RGB image. Therefore, straightforward extensions to the multi/hyperspectral based on methods established in computer vision may not be suitable. To the best of our knowledge, there is currently no self-supervised pretext task designed for spectral-spatial information extraction in the context of remote sensing analysis. In this paper, to address the limitations and deal with the spectral and spatial features of remote sensing data, we propose a novel SSL framework to capture the spectral-spatial pattern from massive unlabelled remote sensing data.

\section{The proposed method}

The proposed SSL aims to learn useful high-level representation that keeps both spatial and spectral information from the label-free remote sensing data and demonstrates this representation can be used to boost downstream remote-sensing tasks. The proposed SSL can be used for remote sensing analysis at both object- and pixel- levels:

\begin{enumerate}
    \item One is a object-based SSL (ObjSSL), which is a kind of contrastive learning. This method is suitable for extracting features from high to very high spatial resolution remote sensing data. The ObjSSL proposes a joint spatial-spectral aware multi-view pretext task, which is a classification problem. It uses cross-entropy loss to measure how well the network can classify the representation amongst a set of multi-views of one target. 
    \item The other is a pixel-based SSL (PixSSL), which is a kind of generative learning suitable for low to medium spatial resolution images. We propose a spectral aware pretext task for reconstructing the original spectral profile. A spectral masked, auto encoder-decoder is designed to learn meaningful latent representations.
\end{enumerate}

\begin{figure}[h]
    \centering
    \includegraphics[width=0.45\textwidth]{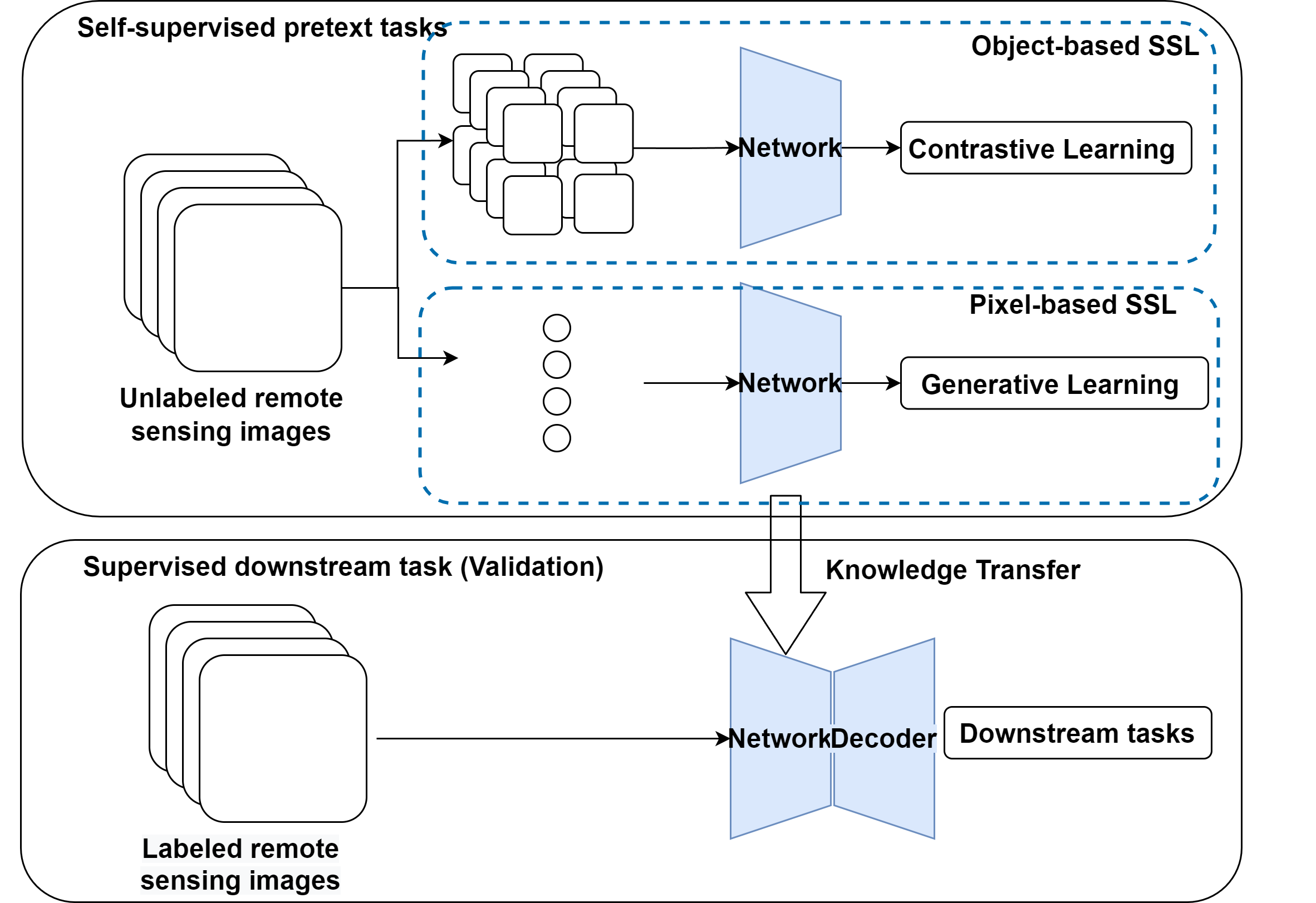}
    \caption{The framework of proposed SSL}
    \label{FIG:1}
\end{figure}

The framework of the proposed SSL method is shown in {Fig.~\ref{FIG:1}}. The first part is the SSL training with unlabelled data, then the trained representations and network is used for downstream tasks through the knowledge transfer. Specific decoder is added after the network, for each specific task. We evaluate the performance of the SSL by using specific downstream tasks (a multi-label land cover classification task on Sentienl-2 multispectral datasets and a ground soil parameter retrieval task on hyperspectral datasets).

\subsection{Object-based SSL (ObjSSL)}

\begin{figure}[h]
    \centering
    \includegraphics[width=0.45\textwidth]{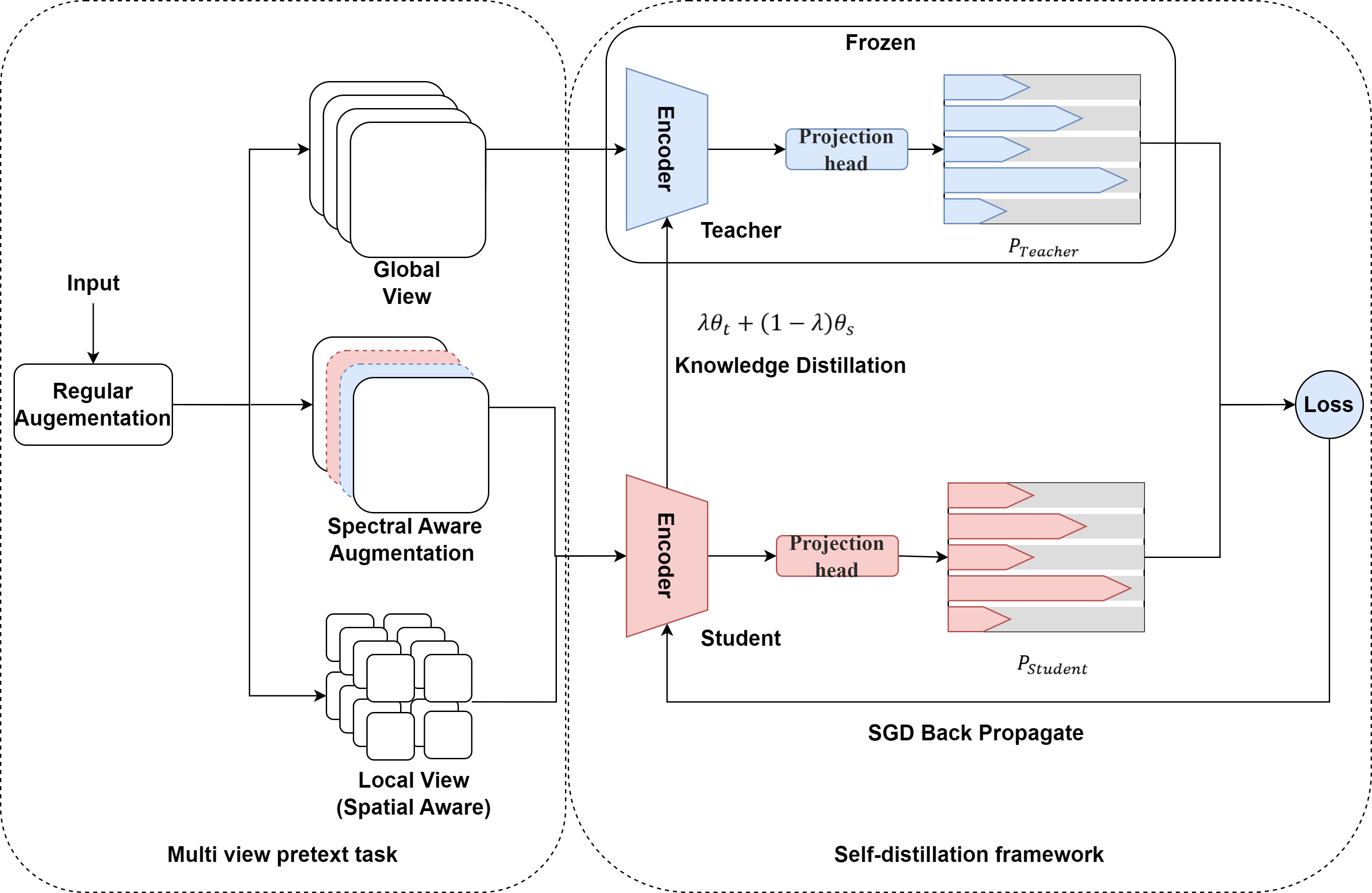}
    \caption{ObjSSL’s architecture.}
    \label{FIG:2}
\end{figure}

The ObjSSL is a contrastive learning method to learn the representation of the remote sensing data. The idea of contrastive learning is to learn representations that bring similar data points (Positive pairs) closer while pushing randomly selected points further away or to maximize contrastive-based mutual information lower bound between different views (Negative pairs). The pretext task of the ObjSSL is a classification problem that uses contrastive loss to measure how well the model can classify the representation among a set of unrelated negative and positive samples. In this work, the positive samples are generated by discerning the representation of augmented views of the same data.  The Negative pairs assume that different images in a batch during model training represent different categories. The flowchart of the work can be shown in {Fig.~\ref{FIG:2}}. There are two main parts in the ObjSSL:
\begin{enumerate}
    \item A novel multi-view pretext task that generates positive pairs for the ObjSSL by generating different views of remote sensing data from both spectral and spatial perspectives. It is a composition of multiple data augmentation operations, including spectral aware augmentation, regular augmentation, and local and global augmentation.
    \item A self-distillation framework that uses two networks to learn the representation from multi-views of the data: the student network and the teacher network. The student network is trained to match the output of a given teacher network.
\end{enumerate}

\subsubsection{Multi-view pretext task}

In ObjSSL, the positive pairs are generated by applying data augmentation to create noise versions of the original samples. An appropriate data augmentation is essential for learning good, generalizable embedding features. It introduces unnecessary changes to the original images without modifying the semantics meaning, thus encouraging the model to learn the essential features.  In \cite{chen_simple_2020}, the author has demonstrated that the composition of multiple data augmentation operations is crucial in defining the contrastive prediction tasks that yield effective representations. In this work, the joint spatial-spectral aware multi-view pretext task is proposed to generate positive pairs of data for the ObjSSL. It consists of a composition of multiple data augmentation operations, including 1) regular augmentation, 2) local and global spatial augmentation, and 3) spectral aware augmentation. 

\paragraph{Regular augmentation} Regular augmentation includes common data transformations, such as: random rotation and zooming, gaussian blur, and random noise. 

\paragraph{Local and global augmentation} Local and global augmentation is used to generate views of different spatial areas. With an input data X of size $120^2$. The output of this augmentation is a set containing global views and several local views of smaller resolutions. We assume that the original data contains the global context. The small crops are called local views that use an image size of $36^2$. It covers less than 50\% of the global view. We assume that it contains the local context. Then two views are fed into the Self-distillation framework. All local views are passed through the student while only the global view is passed through the teacher. It encourages the student network to interpolate context from a small crop and the teacher network to interpolate context from a bigger image. 

\paragraph{Spectral aware augmentation} Spectral aware augmentation is data transformation that performs in parallel with local and global augmentation. The traditional color-based augmentation method is a set of random transformations on random channels, including variations between channels, which inevitably change the spectral order their relative positions. In this work, the spectral-aware augmentation drops the random channels (30\%-50\%) and replaces them with a value of zero. It guarantees that the relationship and relative position of the different channels does not change. This view is passed through the student encoder. It encourages the student network to learn the full spectral context from the teacher network.

\subsubsection{Self-distillation framework}

The self-distillation framework consists of teacher and student networks (encoder), which has the same structure with different parameters ($\theta_t$ and $\theta_s$). In this work, the Spectral–Spatial Vision Transformer \cite{zhang_self-supervised_2022} designed to extract spectral and spatial features from remote sensing data is selected as the encoder.  From a given image $X$, we generate a set of different views ($X_1$,$X_2$,$X_3$….) by the data augmentation. The $X_1$ and  $X_2$ is fed into the teacher and student encoder separately and the outputs are the probability distributions $P_t$ and $P_s$. This can be formulated as:

\begin{equation}
    P_t(x)=\operatorname{SoftMax}\left(g_{\theta_t}(x) / \tau_t\right)
\end{equation}

Where $g$ is the encoder with parameters $\theta$. $\tau$ is a temperature parameter that controls the sharpness of the output distribution. We learn to match these distributions by minimizing the cross-entropy loss between $P_t$ and $P_s$. 

\begin{equation}
    Loss =\min _{\theta_s}\left(-P_t\left(x_1\right) \log P_s\left(x_2\right)\right)
\end{equation}

\begin{figure}[h]
    \centering
    \includegraphics[width=0.45\textwidth]{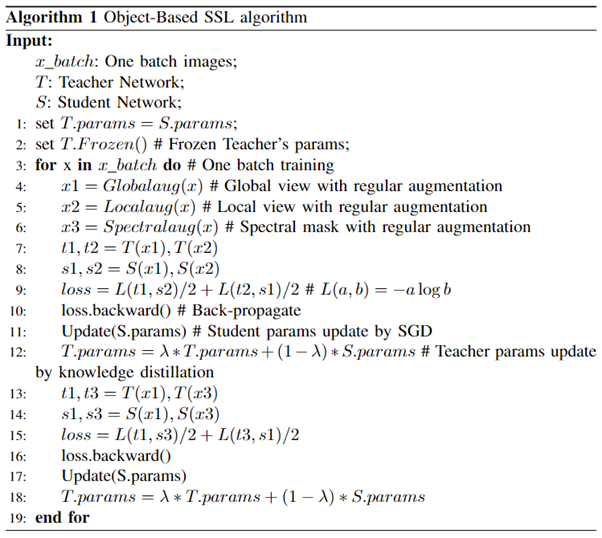}
    \caption{The algorithm of the ObjSSL.}
    \label{FIG:3}
\end{figure}

In this work, the teacher is a momentum teacher, which means that the students' weights $texttheta_S$) are an exponentially moving average. The update rule for the teacher’s weights ($texttheta_t$) is:

\begin{equation}
    \theta_t \leftarrow \lambda \theta_t+(1-\lambda) \theta_s
\end{equation}

with $\lambda$ following a cosine schedule from 0.96 to 1 during training. The algorithm can be summarized in {Fig.~\ref{FIG:3}}. 

\subsection{Pixel-based SSL (PixSSL)}

\begin{figure}[h]
    \centering
    \includegraphics[width=0.45\textwidth]{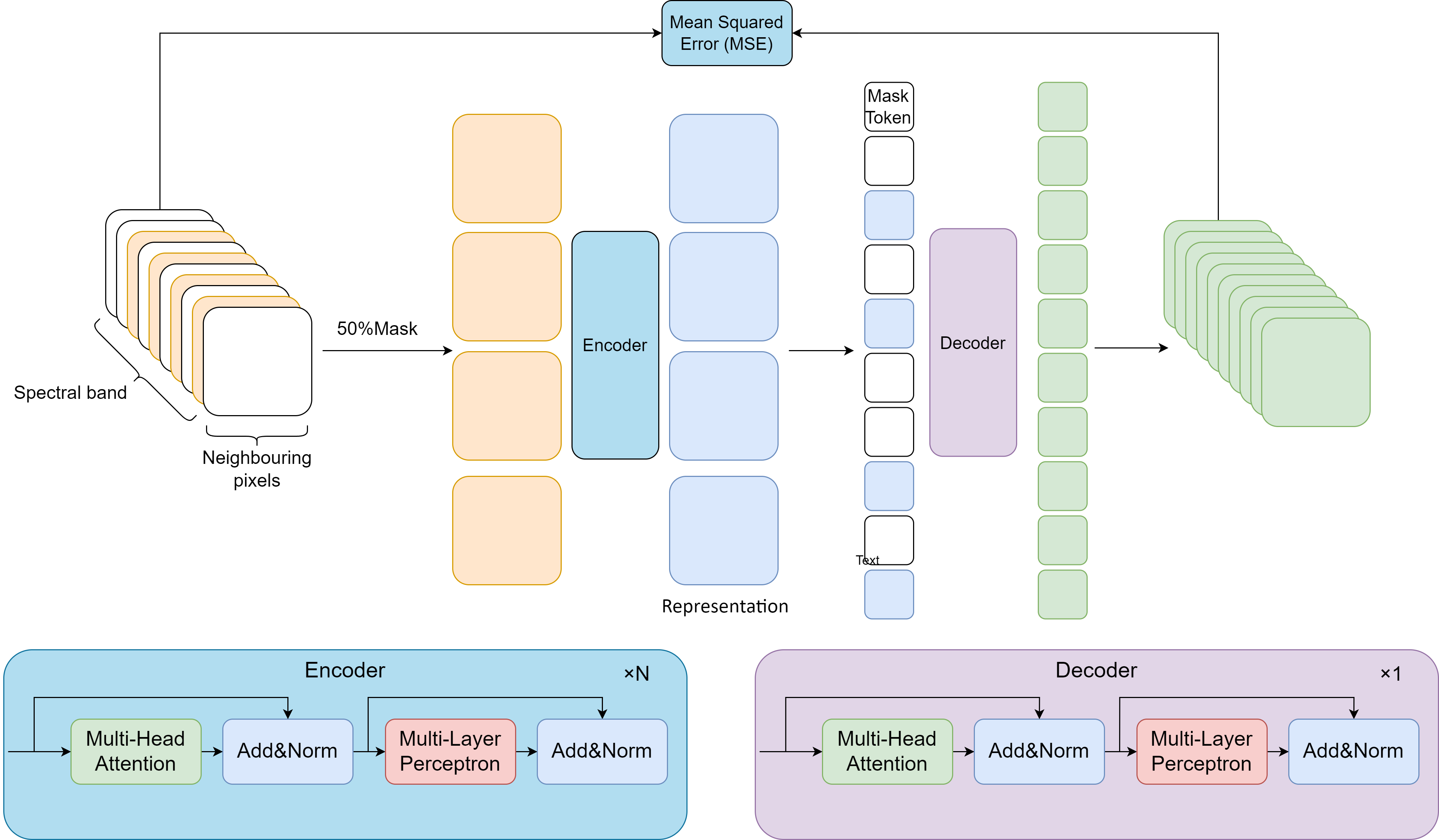}
    \caption{PixSSL’s architecture.}
    \label{FIG:4}
\end{figure}

The PixSSL is a generative SSL method, in which the pretext task is to reconstruct the original input while learning meaningful latent representation. {Fig.~\ref{FIG:4}} shows the architecture of PixSSL. In this work, from a pixel view, we designed a spectral information reconstruction task to learn latent representation from the rich spectral information of remote sensing data. There are three main innovations in the PixSSL:

\begin{enumerate}
    \item To ensure the relationships and relative positions of the different spectral channels remain unchanged, a spectral reconstructive pretext task is introduced to recover each pixel’s spectral profile from masked data. Based on our experiments, we find that masking 50\% of the spectral information yields a meaningful self-supervisory task.
    \item An encoder-decoder architecture is designed to perform this pretext task.  The encoder is used to generate meaningful latent representation and decoder is used to recover the masked spectral profile.
    \item Pixel-based analysis methods require processing every pixel within an image, which significantly increases the amount of computation. To optimise computational efficiency, our proposed encoder can operate on a subset of the spectral data (masked data) to reduce the data input. Meanwhile, the aim of the SSL is to train an encoder to generate meaningful latent representation for downstream tasks. Therefore, we only added a lightweight decoder that reconstructs the spectral profile to reduce computational consumption.
\end{enumerate}

\begin{figure}[h]
    \centering
    \includegraphics[width=0.45\textwidth]{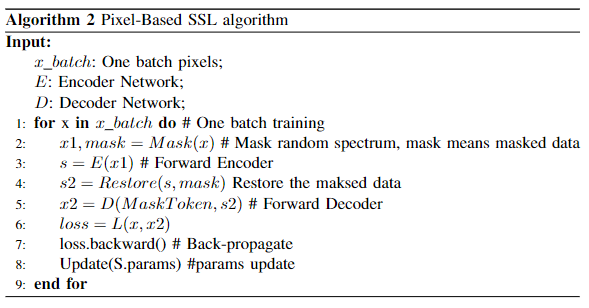}
    \caption{The algorithm of the PixSSL}
    \label{FIG:5}
\end{figure}

The algorithm can be summarized in {Fig.~\ref{FIG:5}}.  

\subsubsection{Spectral reconstructive pretext task}

The self-supervised pretext task in PixSSL aims to recover spectral information from masked data. In this work, we use high masking ratios to randomly mask each data’s spectral profile. The high ratios largely eliminate redundancy, resulting in a pretext task that cannot be easily solved by extrapolation from visible neighboring bands. Through the PixSSL performance experiment, we have demonstrated that masking 50\% of the spectral information yields a meaningful latent representation.

\subsubsection{The spectral masked auto encoder-decoder network}

In this work, we have proposed a spectral-masked autoencoder that reconstructs the original spectral information given its partial spectral. Our approach has an encoder that maps the pixel’s spectral information to a latent representation, and a decoder that reconstructs the spectral profile from the latent representation. {Fig.~\ref{FIG:4}} illustrates the flowchart of the PixSSL. We use an asymmetric design that allows the encoder to operate on masked partial spectral, and a decoder that reconstructs the full spectral from the latent representation and mask tokens. The last layer of the decoder is a linear projection whose number of outputs equals the number of the spectral channel of the data. The loss function computes the mean squared error (MSE) between the reconstructed and original data in the pixel space. 

\paragraph{Encoder}

The encoder in this work is a transformer encoder but applied only on unmasked data. Only 50\% of the spectral channel is used for the encoder in the SSL training. Our encoder embeds patches by a linear projection with added positional embeddings and then processes the resulting set via a number (N) of Encoder blocks. 

There are four main parts in the encoder as shown in {Fig.~\ref{FIG:4}}: Multi-Head Self Attention layer (MSP), Multi-Layer Perceptrons (MLP), Layer Norm, and Residual connections, which were introduced in CNN evolution \cite{he_deep_2015}.

\textbf{Multi-Head Self Attention layer (MSP)}

\begin{figure}[h]
    \centering
    \includegraphics[width=0.46\textwidth]{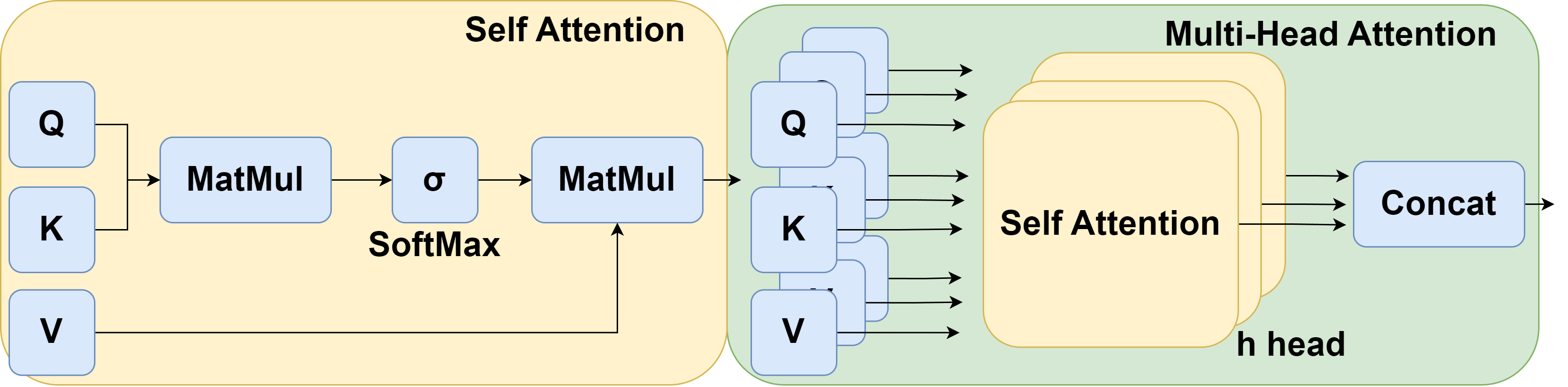}
    \caption{The architecture of MSP}
    \label{FIG:6}
\end{figure}

The MSP is the core of the transformer, and it consists of several self-attention blocks (h) that integrate multiple complicated interactions between different elements in the sequence ({Fig.~\ref{FIG:6}}). The self-attention mechanism can perform the non-local operation, capturing long-range dependencies/global information between selected patches in the sMRI image \cite{buades_non-local_2005}. Here, we denote the input of the model as a sequence of n patches ($p_1$,$p_2$…$p_n$) by  $\mathbf{P} \in \mathrm{R}^{n \times d}$, where $d$ is the embedding dimension of each patch. The goal of self-attention is to capture the interaction amongst all n patches by encoding each patch in terms of the global contextual information, which is done by defining three learnable weight matrices to transform Queries ($\mathrm{W}^Q \in \mathrm{R}^{d \times d_q}$), Keys ($\mathrm{W}^K \in \mathrm{R}^{d \times d_k}$) and Values ($\mathrm{W}^V \in \mathrm{R}^{d \times d_v}$), where $d_q=d_k$. The input $P$ is first projected into Queries (Q), Keys (K) and Values (V) by using 1×1×1convolution filter which can be defined as:

\begin{equation}
    Q=P \mathrm{~W}^Q
\end{equation}
\begin{equation}
    K=P \mathrm{~W}^K
\end{equation}
\begin{equation}
    V=P \mathrm{~W}^V
\end{equation}
The output of the self-attention layer is:

\begin{equation}
 A=\operatorname{softmax}\left(\frac{Q K^T}{\sqrt{d_q}}\right) V   
\end{equation}

The self-attention computes the dot-product of the query with all keys, which is then normalized using the SoftMax operator to get the attention scores. Each patch becomes the weighted sum of all patches in the image, where the attention scores give weights.

Then, each self-attention block has its own learnable weight ($W^{Q_i}$,$W^{K_i}$,$W^{V_i}$,$i \in h$). The output of the h self-attention blocks ($A_i$) in multi-head attention is then concatenated into a single matrix and then subsequently projected to another weight matrix $W^mh$. The operation is shown in {Fig.~\ref{FIG:6}} and can be formulated as:

\begin{equation}
    M S P=\sum_{i=1}^h A_i \times W^{m h}
\end{equation}

Then, for building a deeper model, a residual connection is employed around each module, followed by Layer Normalization \cite{ba_layer_2016}. Layer Norm is the normalization method in the NLP area instead of Batch norm in vision tasks. It is applied before every block as it does not introduce any new dependencies between the training images. It helps to improve training time and generalization performance. The operation can be written as:

\begin{equation}
S=\operatorname{LayerNorm}(M S P(P)+P)
\end{equation}

\textbf{Multi-Layer Perceptrons (MLP)}

An MLP is a particular case of a feedforward neural network where every layer is a fully connected layer. An MLP is added at the end of each MRI transformer block, containing two fully connected layers (Fc1 and Fc2) with Gaussian Error Linear Unit (GELU). It has been proven to be an essential part of the transformer that stops and drastically slows down rank collapse in model training \cite{dong_attention_2021}. Residual connections are applied after every block as they allow the gradients to flow through the network directly without passing through nonlinear activations. The output of MLP can be written as:

\begin{equation}
F=\operatorname{LayerNorm}(S+F c 2(\operatorname{GELU}(F c 1(S)))
\end{equation}

\paragraph{Decoder}

The input to the decoder is the full set of tokens consisting of (i) encoded visible patches, and (ii) mask tokens. Each mask token is a shared, learned vector that indicates the presence of a missing patch to be predicted. We add positional embeddings to all tokens in this full set; without this, mask tokens would have no information about their location in the image. The decoder has another series of Transformer blocks. The decoder is only used during pre-training to perform the image reconstruction task (only the encoder is used to produce image representations for recognition).

\section{Experiments evaluation}

This section is devoted to illustrating the capabilities of the presented algorithm in two typical application scenarios and data. There are two main experiments: in the first experiment we evaluate the performance of ObjSSL through a downstream multi-label classification task. The most common medium-resolution multispectral Sentinel 2 is selected as the data source.  In the second experitment, we measure the performance of PixSSL using hyperspectral data through a soil parametric regression task.

\subsection{ObjSSL performance evaluation}

In this work, we evaluate the performance of the proposed ObjSSL through a downstream multi-label classification task. We have conducted three types of experiments: 
\begin{enumerate}
    \item Sensitivity Analysis of the Proposed Approach. In this experiment, we perform the sensitivity analysis of the proposed approach under different settings and strategies. Firstly, we analyze the downstream task performance with and without the spectral aware augmentation to evaluate the impact of the designed pretext task. Then we report the model performance with 5\%, 25\%, 50\%, and 100\% of the training data with and without the SSL to demonstrate the effect of SSL on the supervised classification task.

    \item Comparison among the existing SSL methods. A comparative experiment is proposed to report the accuracy of this proposed algorithm with the three latest contrastive learning SSL methods including MoCo-V2 \cite{chen_exploring_2020},  BYOL \cite{grill_bootstrap_2020}, and DINO \cite{caron_emerging_2021}.

    \item Comparison among the existing backbones. A comparative experiment is proposed to report the accuracy of this proposed algorithm with the three commonly used deep learning classification networks including VGG 16 \cite{simonyan_very_2014}, ResNet 50 \cite{he_deep_2015}, and Vision transformer \cite{dosovitskiy_image_2020}.
\end{enumerate}

\subsubsection{Data Collection}

A public dataset BigEarthNet \cite{sumbul_bigearthnet-mm_2021} is selected for this experiment. 125 Sentinel-2 tiles acquired between June 2017 and May 2018 from the 10 countries (Austria, Belgium, Finland, Ireland, Kosovo, Lithuania, Luxembourg, Portugal, Serbia, and Switzerland) of Europe are initially selected. All the tiles are atmospherically corrected by the Sentinel-2 Level 2A product generation and formatting tool (sen2cor). Then, each tile is divided into non-overlapping image patches with the size of 120*120. Each image patch was annotated by the multiple land-cover classes (i.e., multi-labels) that are provided by the CORINE Land Cover database of the year 2018 (CLC 2018). The CLC Level-3 nomenclature is interpreted and arranged in a new nomenclature of 19 classes (see {Table.~\ref{table:2}}). Ten classes of the original CLC nomenclature are maintained in the new nomenclature, 22 classes are grouped into 9 new classes, and 11 classes are removed. There are a total of 519,284 patches of data. Since the data may have been acquired in the same geographical area at different times. The result may not be reliable due to the possibility of the data acquired in the same place appearing in both the training and prediction sets. To avoid this issue, the train and validation sets do not share the images acquired in the same geographical area. As a result, we use the data list in \cite{sumbul_bigearthnet_2020,sumbul_deep_2020}. {Table.~\ref{table:2}} shows the number of images of each class associated with training and validation sets.

\begin{table}[h]
\caption{Number of images of each class.}\label{table:2}
\centering
\resizebox{0.49\textwidth}{!}{%
\begin{tabular}{lrr}
\hline
\textbf{Class-Nomenclature}                                                                       & \multicolumn{1}{l}{\textbf{Train}} & \multicolumn{1}{l}{\textbf{Validation}} \\ \hline
\textbf{Urban fabric}                                                                             & 56,963                             & 17,928                                  \\
\textbf{Industrial or commercial units}                                                           & 9,057                              & 2,808                                   \\
\textbf{Arable land}                                                                              & 146,998                            & 47,150                                  \\
\textbf{Permanent crops}                                                                          & 22,538                             & 6,812                                   \\
\textbf{Pastures}                                                                                 & 74,827                             & 24,170                                  \\
\textbf{Complex cultivation patterns}                                                             & 78,565                             & 25,638                                  \\
\textbf{Land principally occupied by   agriculture, with significant areas of natural vegetation} & 98,585                             & 32,052                                  \\
\textbf{Agro-forestry areas}                                                                      & 23,388                             & 7,261                                   \\
\textbf{Broad-leaved forest}                                                                      & 107,170                            & 34,130                                  \\
\textbf{Coniferous forest}                                                                        & 125,243                            & 39,532                                  \\
\textbf{Mixed forest}                                                                             & 133,926                            & 42,641                                  \\
\textbf{Natural grassland and sparsely   vegetated areas}                                         & 9,223                              & 2,799                                   \\
\textbf{Moors, heathland and   sclerophyllous vegetation}                                         & 12,408                             & 3,859                                   \\
\textbf{Transitional woodland, shrub}                                                             & 112,739                            & 36,211                                  \\
\textbf{Beaches, dunes, sands}                                                                    & 1,315                              & 221                                     \\
\textbf{Inland wetlands}                                                                          & 16,751                             & 5,349                                   \\
\textbf{Coastal wetlands}                                                                         & 1,256                              & 310                                     \\
\textbf{Inland waters}                                                                            & 51,100                             & 16,177                                  \\
\textbf{Marine waters}                                                                            & 56,854                             & 18,023                                  \\
\textbf{Total number of images}                                                                   & 393,418                            & 125,866                                 \\ \hline
\end{tabular}

}
\end{table}

\subsubsection{Evaluation Metrics}

The performance evaluation of multi-label classification method requires the analysis of several factors, not just the assessment of the number of correct predictions, and therefore requires complex analysis than in the single-label case. In this work, various classification-based metrics and ranking-based metrics with varying characteristics are selected to accurately evaluate the accuracy of the proposed approach.
Under the category of classification-based metrics, results of experiments were provided in terms of three performance metrics:  1) Accuracy, 2) Precision, 3) Recall, 4) F1 Score and 5) Hamming loss (HL). These metrics are calculated as follows:

\begin{equation}
\text { Accuracy }=\frac{t p+t n}{t p+t n+f p+f n}
\end{equation}

\begin{equation}
\text { Precision }=\frac{t p}{t p+f p}
\end{equation}

\begin{equation}
\text { Recall }=\frac{t p}{t p+f n}
\end{equation}

Where TP, FP, FN and TN indicate the conditions of true positive, false positive, false negative and true negative, respectively. We use macro average for the overall Precision. Recall and F1-score. A macro-average will compute the metric independently for each class and then take the average, which is preferable when the class data imbalance. 

The F1 -Score is the weighted harmonic mean of the correct prediction rates among the considered ground reference and the multi-label predictions.

\begin{equation}
F 1_{\text {score }}=\frac{\text { Recall } \times \text { Precision }}{\text { Recall }+ \text { Precision }} \times 2
\end{equation}

The Hamming loss (HL) is the average Hamming distance between the ground reference labels and predicted multi-labels. Thus, it is defined as follows:

\begin{equation}
HL=\frac{1}{n_{\text {labels }}} \sum_{j=0}^{n_{\text {labels }}-1} 1\left(\hat{y}_j \neq y_j\right)
\end{equation}

Where $\hat{y}_j$ is the predicted value for the $j^{\text {th }}$ label of a given sample, $y_j$ is the corresponding true value, and $n_{\text {labels }}$ is the number of classes or labels. 
Under the category of ranking-based metrics, results of experiments are provided in terms of three performance evaluation metrics: 1) Ranking loss (RL); 2) Coverage (COV); and 3) Label ranking average precision (LRAP). All the ranking-based metrics are defined with respect to the ranking of the $j^{\text {th}}$ label in the class probabilities result of an multi-label classification approach for the $i^{\text {th}}$  image that is defined as $\operatorname{rank}_{i j}=\left|k: P\left(l_k \mid \boldsymbol{x}_i\right) \geq P\left(l_j \mid \boldsymbol{x}_i\right)\right|$. Unlike the classification-based metrics, ranking-based metrics are calculated only by giving equal importance to each sample of the test set.

Accordingly, ranking loss(RL) is the rate of wrongly ordered label pairs (i.e., the probability of a label, which is irrelevant to the image, is higher than a ground reference label), and thus expressed as follows:

\begin{equation}
R L=\frac{1}{M} \sum_{i=1}^M \frac{1}{\left|\boldsymbol{y}_i\right|\left(S-\left|\boldsymbol{y}_i\right|\right)} \sum_{l_j \in \boldsymbol{y}_i} \sum_k \notin \boldsymbol{y}_i r a n k_{i k} \leq \operatorname{rank}_{i j}
\end{equation}

The coverage (COV) calculates the average number of labels required to be included in the prediction list of a multi-label classifier such that all ground reference labels will be predicted. Accordingly, it is defined as follows:

\begin{equation}
C O V=\frac{1}{M} \sum_{i=1}^M \max _{l_j \in y_i} \operatorname{ran}_{i j}
\end{equation}

For each ground reference label, the label ranking average precision (LRAP) calculates the rate of higher-ranked ground reference labels. This is expressed as follows:

\begin{equation}
L R A P=\frac{1}{M} \sum_{i=1}^M \frac{1}{\boldsymbol{y}_i} \sum_{l_j \in \boldsymbol{y}_i} \frac{\left|\left\{l_k: \operatorname{rank}_{i k} \leq \operatorname{rank}_{i j}, l_k \in \boldsymbol{y}_i\right\}\right|}{\operatorname{rank}_{i j}}
\end{equation}

It is worth noting that, smaller values of the Hamming loss, ranking loss and coverage indicate better performance of an approach, whereas higher values of the accuracy, precision, recall, F1 -Score and the LRAP are associated to better performance.

\subsubsection{Experimental Setup}

The model training in this work has two main steps. One is the SSL training without labels. Then the second step is supervised training. We can choose either the SSL-generated weights or the default weights as the initial weights.

The SSL training uses the AdamW optimizer \cite{loshchilov_decoupled_2017} and a batch size of 64, distributed over 3 GPUs (GeForce RTX 2080 Ti). The learning rate is linearly ramped up during the first ten epochs as 1e-3. After this warmup, we decay the learning rate with a cosine schedule. The weight decay also follows a cosine scheduled from 0.04 to 0.4. We execute training for 100 epochs. 

For the supervised training. We first transfer the weights learned from SSL training to initialize the model. AdamW optimizer is used for 100 epochs using a cosine decay learning rate scheduler and 20 epochs of linear warm-up. A batch size of 64, a lower initial learning rate of $1 \mathrm{e}-4$, and a weight decay of 0.05 are used for model training.

\subsection{PixSSL performance evaluation}

In this work, we evaluate the performance of the proposed PixSSL through a downstream parameter regression task. The objective of the task is to estimate the soil parameters, specifically, potassium (K), phosphorus pentoxide ($P_2 O_5$), magnesium (Mg), and pH from the hyperspectral images captured over agricultural areas in Poland. 

\subsubsection{Data Collection}

\begin{figure}[h]
    \centering
    \includegraphics[width=0.45\textwidth]{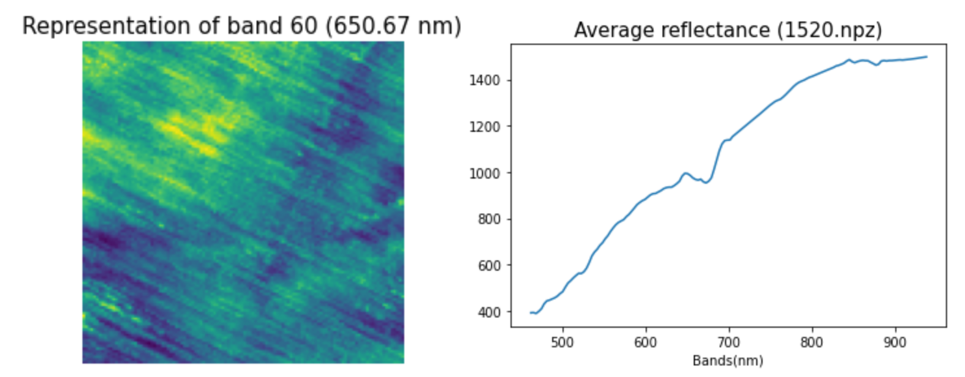}
    \caption{AI4EO hyperspectral data}
    \label{FIG:7}
\end{figure}

In this work, the data from AI4EO hyperspectral challenge is selected \cite{noauthor_ai4eo_nodate}. The dataset comprises 2886 patches in total (2 m GSD), of which 1732 patches are for training and 1154 patches for evaluation. The patch size varies (depending on agricultural parcels) and is on average around 60x60 pixels. Each patch contains 150 contiguous hyperspectral bands (462-942 nm, with a spectral resolution of 3.2 nm). {Fig.~\ref{FIG:7}} shows the data representation of band 60 and the spectral profile of one patch. 

\subsubsection{Experimental Setup}

\begin{figure}[h]
    \centering
    \includegraphics[width=0.45\textwidth]{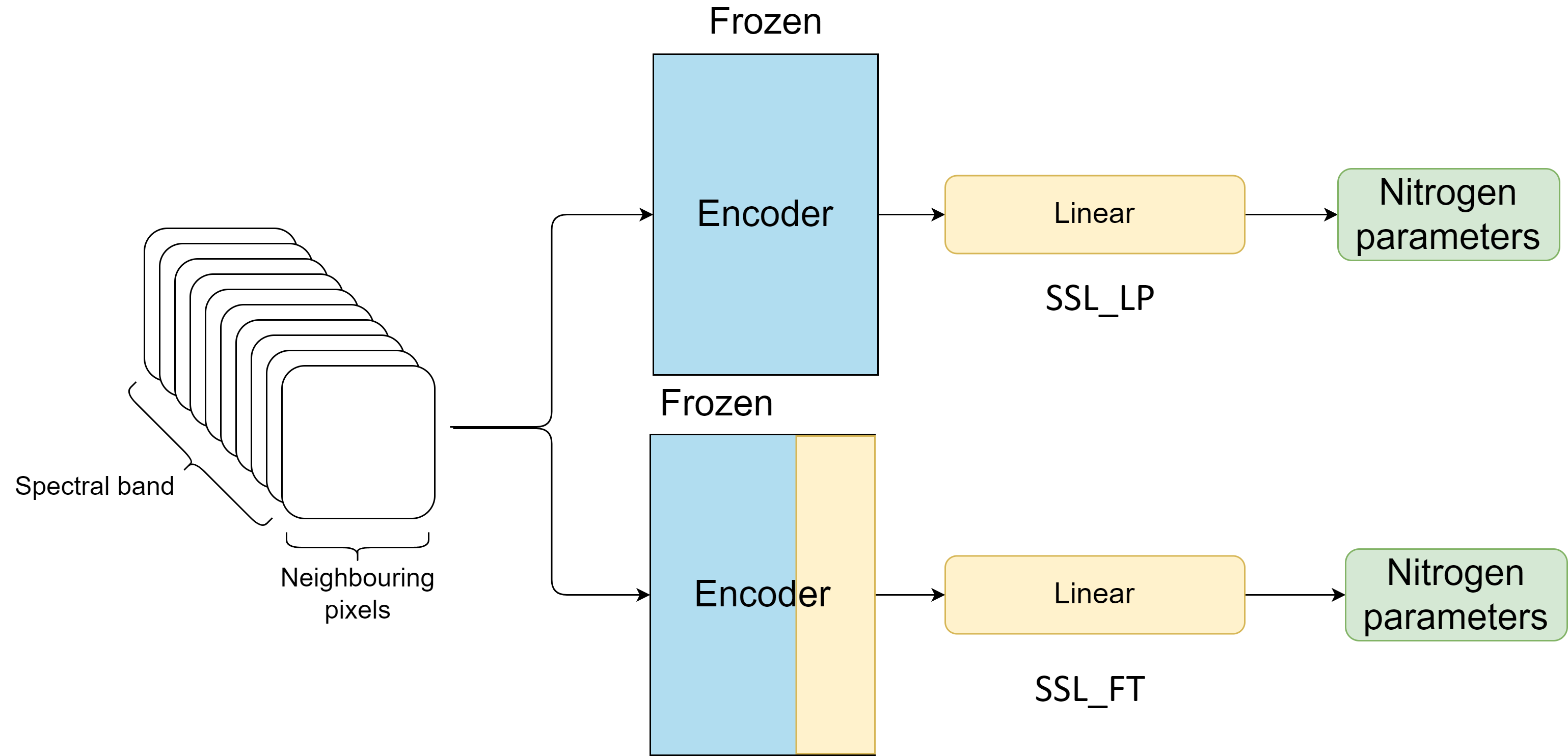}
    \caption{The linear probing (SSL\_LP) and fine-tuning (SSL\_FT) evaluation methods}
    \label{FIG:8}
\end{figure}

Two experiments are proposed to evaluate the performance of the PixSSL. The first one is to determine the best masking ratio. The second is a comparative experiment. Three existing methods are selected for comparison. The baseline method is the machine learning pixel-based method. We assume that each patch is treated as a pixel and average all the values of each waveband in this patch. The spectral profile of each patch is used as the input. The catboost \cite{prokhorenkova_catboost_2019}, as one of the state-of-the-art machine learning regression models, is selected for regression. The root mean squared error (RMSE) and the R-squared (R2) are used to evaluate the model’s performance. In the baseline model, the 1732 spectral profiles extracted from the patch are used for model training.
Then we perform PixSSL on all datasets. We extract around 400 spectral profiles (from 3 x 3 area) from one patch (60x60 pixels). So 1732x400=692,800 are used for the pre-training without labels. At last, we do the downstream regression task to evaluate the representations with two methods: linear probing and fine-tuning ({Fig.~\ref{FIG:8}}).

In linear probing (SSL\_LP), a decoder (linear layer) is stacked on top of the encoder and only the decoder is trained by accessing the labels. Since the encoders have already been trained in the first stage, we freeze all the parameters of the encoder in the downstream task training.

In fine-tuning (SSL\_FT), a similar procedure is followed. In the first stage, encoders are trained without accessing the labels and all the parameters are used as initialization in the second stage. In the second stage a decoder is stacked on top of the backbone, and the whole model is trained by accessing the labels. Notice that we use a smaller learning rate on the encoder to avoid large shifts in weight space.

For the SSL pretraining, we use the AdamW optimizer and a batch size of 512, distributed over 3 GPUs (GeForce RTX 2080 Ti). The learning rate is based on a scaling rule\cite{goyal_accurate_2017}:

\begin{equation}
\mathrm{Lr}=1 \mathrm{e}-4 \times \text { batchsize } / 256
\end{equation}

After this warmup, we decay the learning rate with a cosine schedule. The weight decay also follows a cosine schedule from 0.04 to 0.4. For the supervised training on downstream regression task. We first transfer the weights learned from SSL to initialize the model. AdamW optimizer is used for 100 epochs using a cosine decay learning rate scheduler and 20 epochs of linear warm-up. This learning rate scheduler is only works on decoder and a lower learning rate (1e-6) is used on encoder.

\section{Result}

\subsection{ObjSSL performance}

\subsubsection{Sensitivity Analysis of the Proposed Approach}

In this section, we first evaluate the impact of spectral aware data augmentation on SSL. We report the class-based performance of the proposed model with and without the spectral aware data augmentation ({Table.~\ref{table:3}}).  By analyzing the table, one can see that the model using the spectral aware SSL achieves the highest score for each class compared to the SSL model without spectral aware operation. The average Precision, Recall and F1 scores of the proposed SSL method are 78.66\%, 66.52\% and 71.10\%, which are 5.64\%, 4.40\% and 5.05\% higher than the model without spectral aware operation. This result demonstrates that spectral information in remote sensing data plays a key role in ground object classification.

\begin{table*}[h]
\caption{Class-based performance obtained without and with spectral aware augmentation}\label{table:3}
\centering
\resizebox{0.9\textwidth}{!}{%
\begin{tabular}{lrrrrrr}
\hline
                                                                                         & \multicolumn{3}{c}{W/o spectral aware}                                                    & \multicolumn{3}{c}{With spectral aware}                                                   \\ \hline
                                                                                         & \multicolumn{1}{l}{Precision} & \multicolumn{1}{l}{Recall} & \multicolumn{1}{l}{F1-score} & \multicolumn{1}{l}{Precision} & \multicolumn{1}{l}{Recall} & \multicolumn{1}{l}{F1-score} \\
Urban fabric                                                                             & 72.22\%                       & 65.94\%                    & 68.94\%                      & 86.51\%                       & 71.45\%                    & 78.26\%                      \\
Industrial or commercial units                                                           & 62.34\%                       & 39.79\%                    & 48.58\%                      & 83.62\%                       & 25.39\%                    & 38.95\%                      \\
Mixed forest                                                                             & 82.70\%                       & 78.78\%                    & 80.69\%                      & 85.07\%                       & 82.13\%                    & 83.58\%                      \\
Natural grassland and sparsely vegetated areas                                           & 61.50\%                       & 36.45\%                    & 45.77\%                      & 70.54\%                       & 43.59\%                    & 53.89\%                      \\
Moors, heathland and sclerophyllous vegetation                                           & 71.74\%                       & 46.30\%                    & 56.28\%                      & 73.12\%                       & 61.06\%                    & 66.55\%                      \\
Transitional woodland, shrub                                                             & 66.53\%                       & 61.36\%                    & 63.84\%                      & 69.19\%                       & 66.93\%                    & 68.04\%                      \\
Beaches, dunes, sands                                                                    & 41.43\%                       & 49.15\%                    & 44.96\%                      & 59.72\%                       & 36.44\%                    & 45.26\%                      \\
Inland wetlands                                                                          & 69.05\%                       & 55.14\%                    & 61.31\%                      & 79.48\%                       & 53.54\%                    & 63.98\%                      \\
Coastal wetlands                                                                         & 37.19\%                       & 68.95\%                    & 48.32\%                      & 53.70\%                       & 63.01\%                    & 57.98\%                      \\
Inland waters                                                                            & 88.81\%                       & 79.38\%                    & 83.83\%                      & 90.86\%                       & 80.05\%                    & 85.11\%                      \\
Marine waters                                                                            & 97.43\%                       & 97.77\%                    & 97.60\%                      & 98.25\%                       & 98.06\%                    & 98.15\%                      \\
Arable land                                                                              & 82.88\%                       & 83.74\%                    & 83.31\%                      & 89.03\%                       & 83.66\%                    & 86.26\%                      \\
Permanent crops                                                                          & 71.97\%                       & 38.42\%                    & 50.10\%                      & 73.92\%                       & 60.07\%                    & 66.28\%                      \\
Pastures                                                                                 & 84.83\%                       & 60.43\%                    & 70.58\%                      & 83.39\%                       & 69.27\%                    & 75.67\%                      \\
Complex cultivation patterns                                                             & 73.53\%                       & 57.26\%                    & 64.38\%                      & 74.89\%                       & 66.78\%                    & 70.60\%                      \\
Land principally occupied by agriculture, with significant areas of   natural vegetation & 75.10\%                       & 46.42\%                    & 57.37\%                      & 71.97\%                       & 63.75\%                    & 67.61\%                      \\
Agro-forestry areas                                                                      & 79.29\%                       & 65.04\%                    & 71.46\%                      & 79.59\%                       & 78.61\%                    & 79.10\%                      \\
Broad-leaved forest                                                                      & 84.70\%                       & 62.51\%                    & 71.93\%                      & 81.69\%                       & 76.61\%                    & 79.07\%                      \\
Coniferous forest                                                                        & 84.04\%                       & 87.48\%                    & 85.73\%                      & 90.03\%                       & 83.45\%                    & 86.62\%                      \\
macro avg                                                                                & 73.02\%                       & 62.12\%                    & 66.05\%                      & 78.66\%                       & 66.52\%                    & 71.10\%                      \\ \hline
\end{tabular}

}
\end{table*}

\begin{figure}[h]
    \centering
    \includegraphics[width=0.45\textwidth]{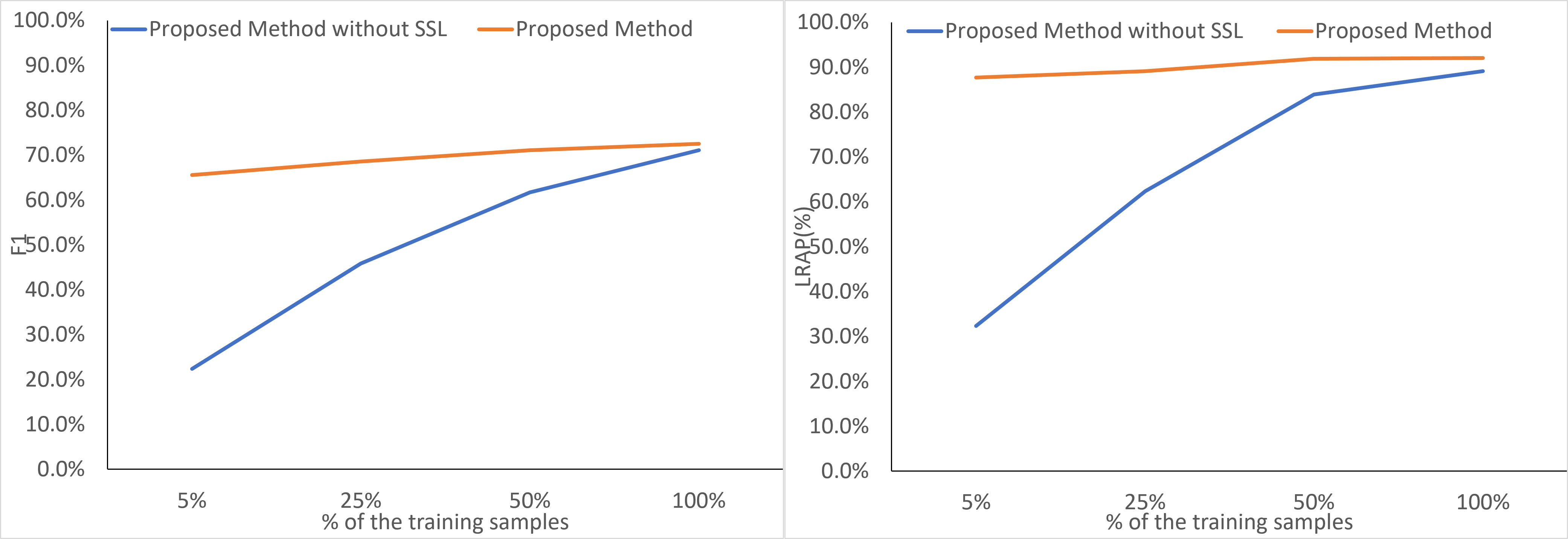}
    \caption{Model performance with 5\%, 25\%, 50\% and 100\% of the training data, reported in terms of F1 scores (\%) and LRAP(\%).}
    \label{FIG:9}
\end{figure}

One of the motivations of SSL is to learn useful representations of data from unlabeled data and then fine-tuning the representations with few labels for the supervised downstream task. In this task, we evaluate the effect of the SSL processing in the downstream task, especially when the amount of training data is limited. {Fig.~\ref{FIG:9}} shows the model performance with 5\%, 25\%, 50\% and 100\% of the training data. The results show that the accuracy of supervised classification on the validation dataset drops significantly when less 50\% of the data is used for training. The F1 score and LRAP is only 22.4\%, 32.4\% when using 5\% of the training data. This indicates that the model is overfitting. When using SSL weights for fine tuning, the model can achieve the best accuracy using only 5\% of the training data.

\subsubsection{Comparison among the existing SSL frameworks}

In the second experiment, we compare the classification results of different SLL frameworks with our proposed method. We pretrain the model with four SSL frameworks (MoCo-V2 \cite{chen_exploring_2020},  BYOL \cite{grill_bootstrap_2020}, DINO \cite{caron_emerging_2021} and proposed SSL) and then fine tune the representations on 50\% of the training data. {Table.~\ref{table:4}} shows the model performance, the model performance without SSL is added as a reference. The results show that the use of SSL can be beneficial in allowing the model to converge on limited data. Compared to the other three SSL frameworks, the proposed SSL method emphasizes the spectral and spatial features in remote sensing data with higher performance than the other three methods.

\begin{table}[h]
\caption{Results obtained by the MOCO-V2, BYOL, DINO and the proposed SSL pretraining and fine-tune on 50\% of the training data}\label{table:4}
\centering
\resizebox{0.49\textwidth}{!}{%
\begin{tabular}{lllll}
\hline
              & MOCO-V2 & BYOL    & DINO    & Proposed Method \\
Accuracy(\%)  & 91.71\% & 91.86\% & 92.05\% & 92.76\%         \\
Precision(\%) & 65.58\% & 70.62\% & 69.42\% & 73.02\%         \\
Recall(\%)    & 57.80\% & 50.68\% & 55.21\% & 62.12\%         \\
F1(\%)        & 60.96\% & 57.03\% & 60.65\% & 66.05\%         \\
HL            & 0.083   & 0.081   & 0.080   & 0.072           \\
COV           & 4.781   & 4.645   & 4.601   & 4.241           \\
LRAP(\%)      & 86.25\% & 86.33\% & 87.17\% & 89.13\%         \\
RL            & 0.054   & 0.051   & 0.049   & 0.038           \\ \hline
\end{tabular}

}
\end{table}

\subsubsection{Comparison among the existing networks}

{Table.~\ref{table:5}} shows the classification-based and rank-based metrics obtained by the proposed method and the three most popular deep learning networks: VGG16, ResNet50 and ViT. Since the BigEearthNet collects 393,000 training data, it is sufficient for most visual tasks. All the deep learning models achieve satisfactory accuracy on supervised tasks. Duo to the residual connection introduced by ResNet \cite{he_deep_2015}, ResNet obtains a superior accuracy than VGG16, which is demonstrated in most computer vision tasks \cite{wightman_resnet_2021}. ViT\cite{dosovitskiy_image_2020}, as a new computer vision architecture, which utilizes a transformer instead CNN to extract features of the data, achieves accuracy performance close to VGG16. Our method introduces a channel information learning module in VIT, the results show higher performance than ResNet and VIT. A minor improvement in accuracy is also obtained with the addition of SSL in both classifications based and rank-based metrics.

\begin{table}[h]
\caption{Results obtained by the VGG16, ResNet50, ViT and the proposed method with and without SSL}\label{table:5}
\centering
\resizebox{0.49\textwidth}{!}{%
\begin{tabular}{llllll}
\hline
              & VGG16   & ResNet50 & ViT     & Proposed Network   w/o SSL & Proposed Network \\ \hline
\rowcolor[HTML]{F2F2F2} 
Accuracy(\%)  & 90.96\% & 92.32\%  & 89.94\% & 92.83\%                    & 93.84\%          \\
Precision(\%) & 68.59\% & 69.62\%  & 57.91\% & 72.66\%                    & 77.87\%          \\
\rowcolor[HTML]{F2F2F2} 
Recall(\%)    & 40.05\% & 58.04\%  & 34.16\% & 61.52\%                    & 69.25\%          \\
F1(\%)        & 45.85\% & 62.52\%  & 39.00\% & 65.10\%                    & 72.50\%          \\
\rowcolor[HTML]{F2F2F2} 
HL            & 0.090   & 0.077    & 0.101   & 0.071                      & 0.062            \\
COV           & 4.849   & 4.502    & 5.303   & 4.130                      & 3.940            \\
\rowcolor[HTML]{F2F2F2} 
LRAP(\%)      & 83.88\% & 87.67\%  & 80.15\% & 89.91\%                    & 92.07\%          \\
RL            & 0.059   & 0.046    & 0.076   & 0.031                      & 0.028            \\ \hline
\end{tabular}

}
\end{table}

\subsection{PixSSL performance}

\begin{figure}[h]
    \centering
    \includegraphics[width=0.45\textwidth]{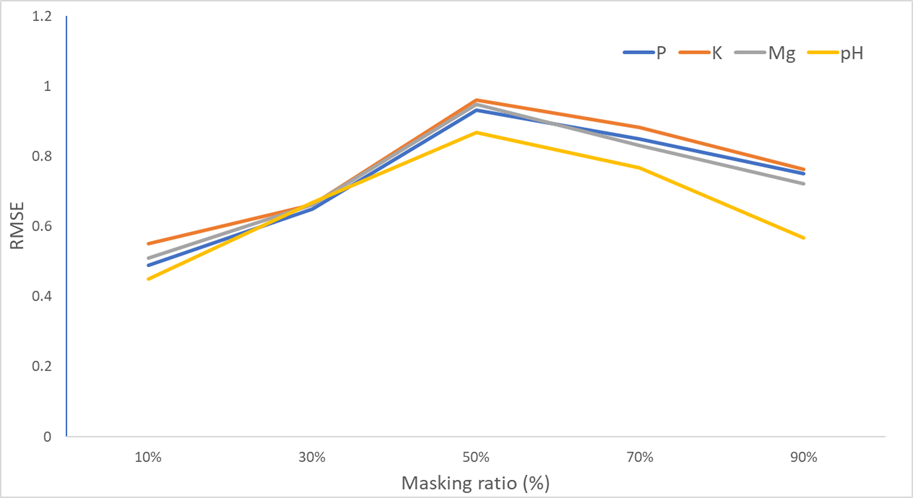}
    \caption{Masking ratio. A high masking ratio (50\% and 70\%) works well for representation learning.}
    \label{FIG:10}
\end{figure}

{Fig.~\ref{FIG:10}} shows the influence of the masking ratio. The optimal ratios are high. The ratio of 50\% is good for the self-supervised representation learning. 

\begin{figure}[h]
    \centering
    \includegraphics[width=0.45\textwidth]{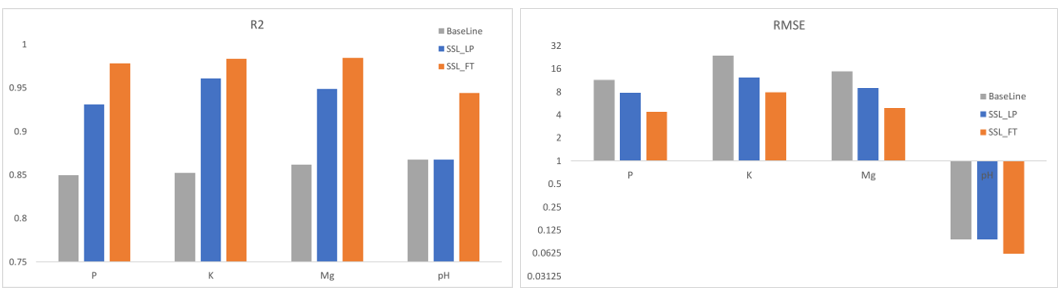}
    \caption{Linear and fine-tune regression result on predicting four nitrogen parameters on the ground target. We report R2 and RMSE accuracy for the evaluations on the validation for the proposed self-supervised method and machine learning-based method.}
    \label{FIG:10}
\end{figure}

In this section, we evaluate the performance of PixSSL on a downstream regression task. Figure 10 shows the R2 and RMSE accuracy on the baseline method and proposed SSL method. With the traditional machine learning pixel-based method, the $R^2$ of nitrogen parameters prediction is around 0.85. With the SSL representation, the $R^2$ of the SSL\_LP increases from 0.93 to 0.95 on P, K, and Mg regression. There is no significant improvement in pH regression since the values of PH on the ground are close. When we fine-tune the final layer of the encoder, the $R^2$ of the proposed model is improved with over 0.95. The result indicates that the representations learned from SSL provide a better prediction of nitrogen properties than using original spectral information only.

\section{Discussion}

In this work, we propose an SSL framework for feature extraction of remote sensing data at both pixel-based and object-based scales. By validating downstream tasks, our results demonstrate that, the new representation of the data learned by the SSL can achieve better performance on downstream tasks than using original data only. In general, the representations learned by the SSL are abstract and cannot be interpreted directly. In the following section, we visualize the representations and discuss their potential value.

\subsection{The representation of ObjSSL}

In the ObjSSL, a novel multi-view pretext task is proposed to generate representation from unlabeled dataset. In our experiments, we demonstrate that our proposed unsupervised learning method exhibits the three main advantages: 1) With the joint spatial-spectral aware pretext task, the deep learning model obtains both spectral and spatial features from the remote sensing data. The classification performance of some spectrum-sensitive categories, such as Mixed Forest, Coniferous Forest, Natural grassland and sparsely vegetated areas, Wetlands and Arable land has been significantly improved. 2) The representations generated from self-supervised learning improve the performance of downstream tasks. 3) After pre-training with self-supervised learning, the deep learning model can be converged faster and better in supervised training with limited dataset. This shows that self-supervised learning generalizes well to the spectral-spatial features in the data.

\begin{figure}[h]
    \centering
    \includegraphics[width=0.45\textwidth]{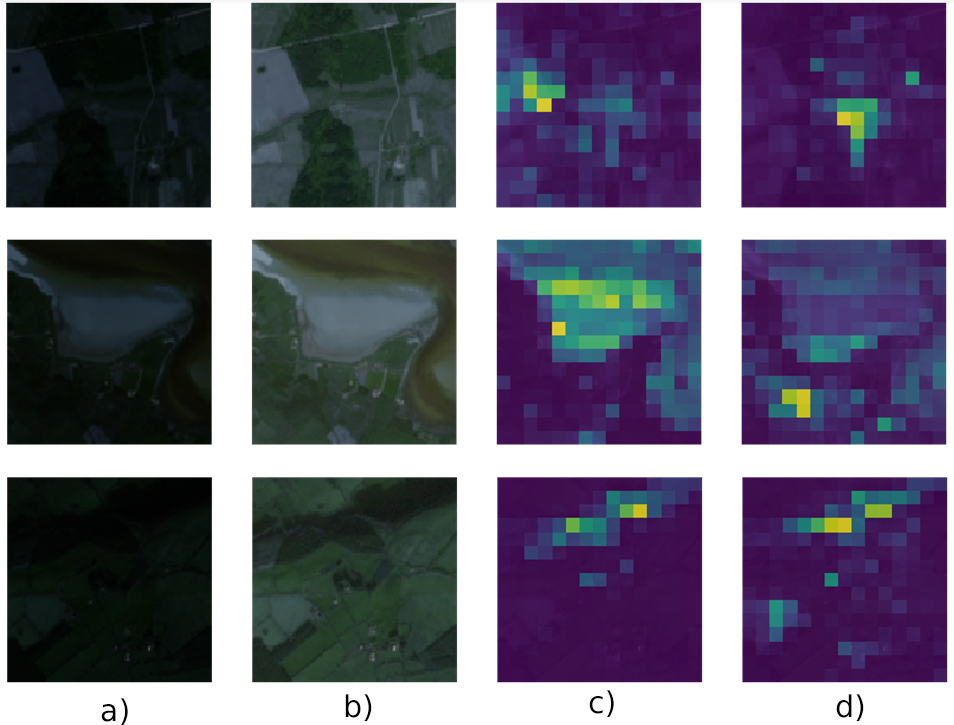}
    \caption{The visualization of the attention map from the last layer in the encoder. a) and b) shows the RGB vision of the data with their enhancements. C) and d) visualize the different attention masp of the last layer in the encoder after ObjSSL.}
    \label{FIG:12}
\end{figure}

In {Fig.~\ref{FIG:12}}, we visualize the attention map for the different heads of the last layer of the encoder after ObjSSL. The a) column is the original data displayed by red, green, and blue channels. In the b) column, we adjust the brightness of the image for better display. The c) and d) visualize the different attention map of the last layer in the encoder after ObjSSL. It shows that the attention map can attend to different semantic regions of an image, which demonstrates that the representations obtained by SSL reflect the semantic information of the data. We believe that this representation has potential to be used in land cover/use tasks.

\begin{figure}[h]
    \centering
    \includegraphics[width=0.45\textwidth]{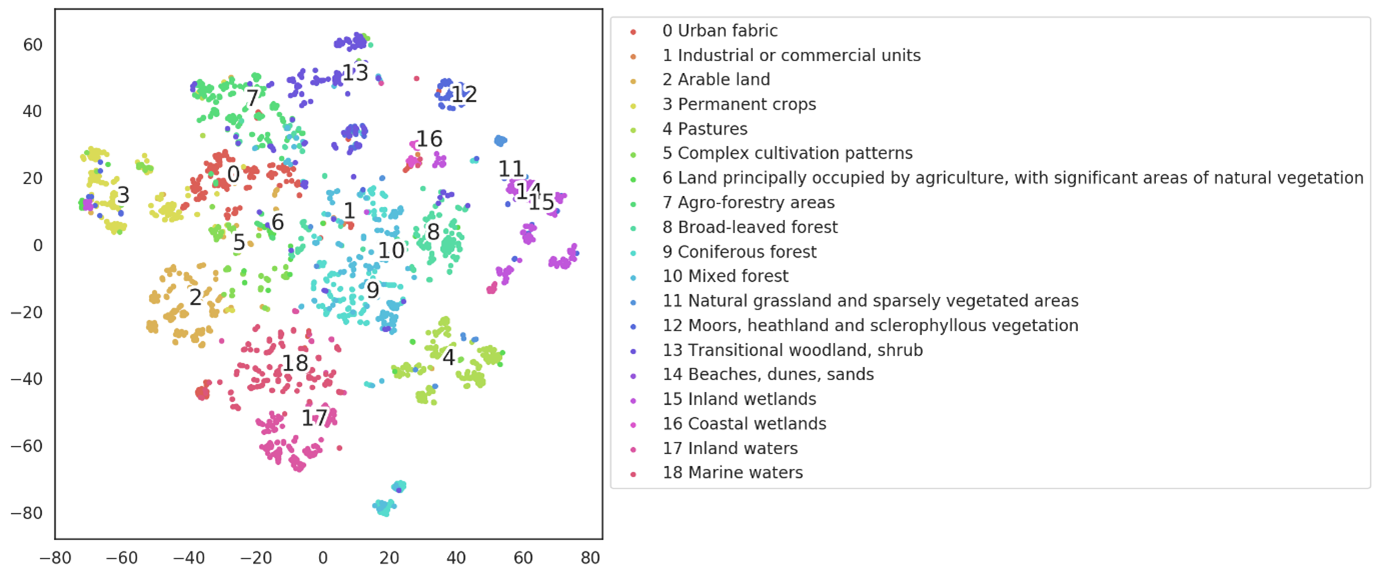}
    \caption{t-SNE visualization of BigEarthNet classes as represented using ObjSSL. For each class, we obtain the embedding by taking the average feature for all images of that class in the validation set.}
    \label{FIG:13}
\end{figure}

In {Fig.~\ref{FIG:13}}, we represent each BigEarthNet class by using the average feature vector for its validation data. We run t-SNE for 5000 iterations and present the resulting class embeddings in Figure 12. The result shows that the representation learned by ObjSSL recovers structures between classes, and similar ground objects are grouped: The water-related classes, such as inland (17) and marine waters (18) are at the bottom. Broad-leaved forests (8), Coniferous forests (9), and mixed forests (10) are grouped in the middle. Natural grassland and sparsely vegetated areas (11), Moors, heathland, and sclerophyllous vegetation (12), Transitional woodland, shrub (13), beaches, dunes, sand (14), and inland wetlands (15) are grouped in the top right. Arable land (2) and permanent crops (3) are on the left.

\subsection{The representation of PixSSL}

\begin{figure}[h]
    \centering
    \includegraphics[width=0.45\textwidth]{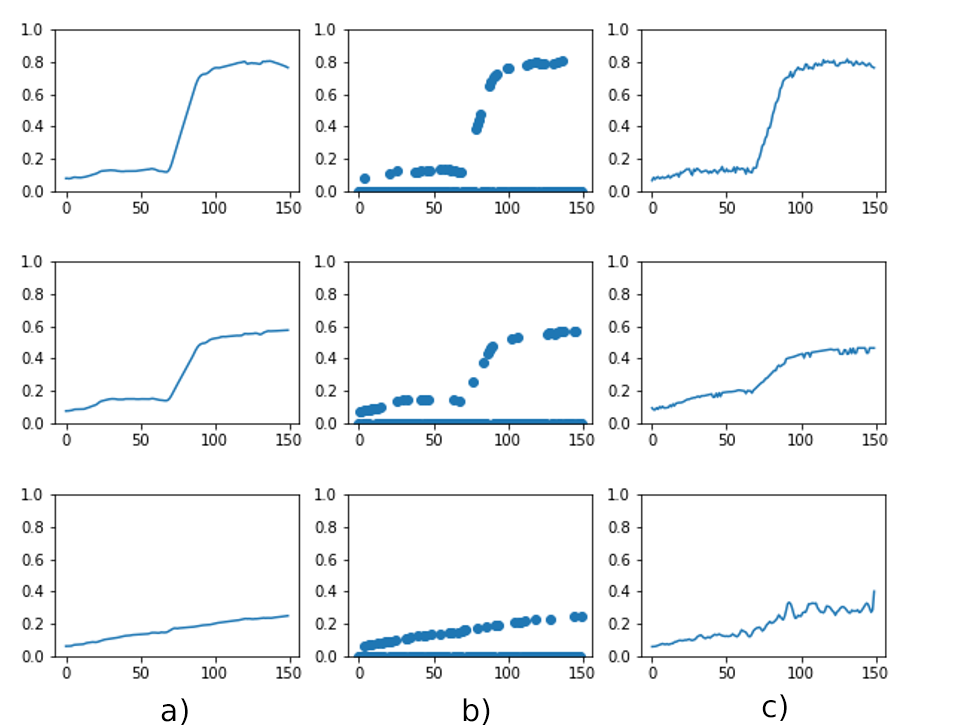}
    \caption{Reconstructions of spectral information using PixSSL. The predictions are reasonably different from the original spectral information but are essentially close, which indicates that the method can be generalized.}
    \label{FIG:14}
\end{figure}

In PixSSL, a reconstruction pretext task is proposed to generate representation from unlabeled data. Our experimental results demonstrate that the representations obtained by SSL can significantly improve the accuracy of pixel-based analysis tasks. In {Fig.~\ref{FIG:14}}, we display how PixSSL reconstructs the masked spectrum. The a) column shows the original spectral profile, and the b) column shows the masked spectral profile where 50\% of the data are masked. The c) column shows the reconstructions of the spectral information. The first two rows show the spectral information of vegetation, and the third row shows bare soil. As can be seen from the results, despite 50\% of the data being masked, our PixSSL recovers the spectral curve well. It can be concluded that this PixSSL learns the spectral features from massive unlabeled data very well.

\subsection{Challenges and future directions}

In this work, we have demonstrated that the SSL learning can enhance the performance of remote sensing applications with remarkable efficiency, by greatly reducing the dependence of deep models on large amounts of annotated data. Nevertheless, as an emerging field within computer vision, it still faces the following hurdles. 1) Computing efficiency. The SSL usually requires significant computational resources due to the large amount of pre-trained data, complex and varied data enhancements, large batch size of training and more training epochs than other existing supervised learning, etc. Meanwhile, with the growth in popularity of hyper NPL model, such as BERT \cite{devlin_bert_2019}, ChatGPT \cite{brown_language_2020}, LaMDA \cite{thoppilan_lamda_2022}, etc. SSL is also widely used to train mega models, which poses a serious challenge to computational resources. At present, little work has been done to reduce computational costs on SSL, but this is an important factor in practical use. The effective data loading, model design, parallel computing and hardware acceleration are therefore to be explored. 2) Prompt Engineering, also known as contextual prompting, refers to methods of how to communicate with large deep learning models to guide their behavior towards desired outcomes without updating their weights \cite{diao_active_2023,liu_what_2021,saravia_prompt_2022}. Self-supervised learning often requires tremendous computational resources to train large models, which poses challenges for non-enterprise researchers. The prompt engineering is an empirical science and does not require large computational resources. The effectiveness of hint engineering methods can vary considerably between models, and therefore requires a lot of experience and experimentation. We believe this will also be a major area of research for SSL in the future.

\section{Conclusion}

In this work, we have proposed a generic self-supervised learning framework based on remote sensing data at both object and pixel levels. This proposed SSL method learns a target representation that covers both spatial and spectral information from massive unlabelled data. This representation has shown to achieve superior performance in downstream remote sensing tasks than using original data as input. More importantly, this approach can alleviate the problem of expensive label of remote sensing data on traditional supervised learning. In this paper, we have designed two experiments with real data. One is land cover classification task based on Sentienl-2 multispectral datasets, we have selected an object-based analysis approach and the results demonstrate that our proposed ObjSSL outperform other traditional SSL methods that are not designed for both spectral-spatial features extraction. The other one is ground soil parameter retrieval tasks on hyperspectral datasets. We have selected a pixel-based analysis method to utilise the rich spectral information. The results demonstrate that the proposed PixSSL can learn improved spectral representations by recovering the spectral information from the masked data. Simultaneously, we visualize the learned representation of the proposed SSL, and the results show that our SSL can learn representation from both spectral-spatial information of unlabelled datasets. We believe that this approach has the potential to be effective in a wider range of remote sensing applications and we will explore its utility in more remote sensing applications in the future.


%

\section*{Acknowledgment}

This work is supported by the BBSRC projects (BB/R019983/1,BB/S020969/1).

\ifCLASSOPTIONcaptionsoff
  \newpage
\fi



\bibliographystyle{IEEEtran}
\bibliography{IEEEabrv,bibtex/bib/IEEEexample}

\begin{thebibliography}{10}
\providecommand{\url}[1]{#1}
\csname url@samestyle\endcsname
\providecommand{\newblock}{\relax}
\providecommand{\bibinfo}[2]{#2}
\providecommand{\BIBentrySTDinterwordspacing}{\spaceskip=0pt\relax}
\providecommand{\BIBentryALTinterwordstretchfactor}{4}
\providecommand{\BIBentryALTinterwordspacing}{\spaceskip=\fontdimen2\font plus
\BIBentryALTinterwordstretchfactor\fontdimen3\font minus
  \fontdimen4\font\relax}
\providecommand{\BIBforeignlanguage}[2]{{%
\expandafter\ifx\csname l@#1\endcsname\relax
\typeout{** WARNING: IEEEtran.bst: No hyphenation pattern has been}%
\typeout{** loaded for the language `#1'. Using the pattern for}%
\typeout{** the default language instead.}%
\else
\language=\csname l@#1\endcsname
\fi
#2}}
\providecommand{\BIBdecl}{\relax}
\BIBdecl

\bibitem{ban2015global}
Y.~Ban, P.~Gong, and C.~Giri, ``Global land cover mapping using earth
  observation satellite data: Recent progresses and challenges,'' pp. 1--6,
  2015.

\bibitem{li_recent_2020}
\BIBentryALTinterwordspacing
D.~Li, P.~Zhang, T.~Chen, and W.~Qin, ``\BIBforeignlanguage{en}{Recent
  {Development} and {Challenges} in {Spectroscopy} and {Machine} {Vision}
  {Technologies} for {Crop} {Nitrogen} {Diagnosis}: {A} {Review}},''
  \emph{\BIBforeignlanguage{en}{Remote Sensing}}, vol.~12, no.~16, p. 2578,
  Jan. 2020, number: 16 Publisher: Multidisciplinary Digital Publishing
  Institute. [Online]. Available:
  \url{https://www.mdpi.com/2072-4292/12/16/2578}
\BIBentrySTDinterwordspacing

\bibitem{osco_review_2021}
\BIBentryALTinterwordspacing
L.~P. Osco, J.~Marcato~Junior, A.~P. Marques~Ramos, L.~A. de~Castro~Jorge,
  S.~N. Fatholahi, J.~de~Andrade~Silva, E.~T. Matsubara, H.~Pistori, W.~N.
  Gonçalves, and J.~Li, ``\BIBforeignlanguage{en}{A review on deep learning in
  {UAV} remote sensing},'' \emph{\BIBforeignlanguage{en}{International Journal
  of Applied Earth Observation and Geoinformation}}, vol. 102, p. 102456, Oct.
  2021. [Online]. Available:
  \url{https://www.sciencedirect.com/science/article/pii/S030324342100163X}
\BIBentrySTDinterwordspacing

\bibitem{ghamisi_advanced_2017}
P.~Ghamisi, J.~Plaza, Y.~Chen, J.~Li, and A.~J. Plaza, ``Advanced {Spectral}
  {Classifiers} for {Hyperspectral} {Images}: {A} review,'' \emph{IEEE
  Geoscience and Remote Sensing Magazine}, vol.~5, no.~1, pp. 8--32, Mar. 2017.

\bibitem{richards_remote_2006}
J.~A. Richards, \emph{Remote {Sensing} {Digital} {Image} {Analysis}}.\hskip 1em
  plus 0.5em minus 0.4em\relax Springer Berlin Heidelberg, 2006.

\bibitem{chen_geographic_2018}
G.~Chen, Q.~Weng, G.~J. Hay, and Y.~He, ``Geographic object-based image
  analysis ({GEOBIA}): {Emerging} trends and future opportunities,''
  \emph{GIScience \& Remote Sensing}, vol.~55, no.~2, pp. 159--182, 2018,
  publisher: Taylor \& Francis.

\bibitem{pal_assessment_2003}
\BIBentryALTinterwordspacing
M.~Pal and P.~M. Mather, ``An assessment of the effectiveness of decision tree
  methods for land cover classification,'' \emph{Remote Sensing of
  Environment}, vol.~86, no.~4, pp. 554--565, Aug. 2003. [Online]. Available:
  \url{http://www.sciencedirect.com/science/article/pii/S0034425703001329}
\BIBentrySTDinterwordspacing

\bibitem{cortes_support-vector_1995}
\BIBentryALTinterwordspacing
C.~Cortes and V.~Vapnik, ``\BIBforeignlanguage{en}{Support-vector networks},''
  \emph{\BIBforeignlanguage{en}{Machine Learning}}, vol.~20, no.~3, pp.
  273--297, Sep. 1995. [Online]. Available:
  \url{http://link.springer.com/10.1007/BF00994018}
\BIBentrySTDinterwordspacing

\bibitem{breiman_random_2001}
\BIBentryALTinterwordspacing
L.~Breiman, ``\BIBforeignlanguage{en}{Random {Forests}},''
  \emph{\BIBforeignlanguage{en}{Machine Learning}}, vol.~45, no.~1, pp. 5--32,
  Oct. 2001. [Online]. Available: \url{https://doi.org/10.1023/A:1010933404324}
\BIBentrySTDinterwordspacing

\bibitem{pal_random_2005}
M.~Pal, ``Random forest classifier for remote sensing classification,''
  \emph{International Journal of Remote Sensing}, vol.~26, no.~1, pp. 217--222,
  2005.

\bibitem{safari_comparative_2017}
A.~Safari, H.~Sohrabi, S.~Powell, and S.~Shataee, ``A comparative assessment of
  multi-temporal {Landsat} 8 and machine learning algorithms for estimating
  aboveground carbon stock in coppice oak forests,'' \emph{International
  Journal of Remote Sensing}, vol.~38, no.~22, pp. 6407--6432, 2017, publisher:
  Taylor \& Francis.

\bibitem{singh_remote_2022}
C.~Singh, S.~K. Karan, P.~Sardar, and S.~R. Samadder, ``Remote sensing-based
  biomass estimation of dry deciduous tropical forest using machine learning
  and ensemble analysis,'' \emph{Journal of Environmental Management}, vol.
  308, p. 114639, 2022, publisher: Elsevier.

\bibitem{garcia-garcia_review_2017}
\BIBentryALTinterwordspacing
A.~Garcia-Garcia, S.~Orts-Escolano, S.~Oprea, V.~Villena-Martinez, and
  J.~Garcia-Rodriguez, ``A {Review} on {Deep} {Learning} {Techniques} {Applied}
  to {Semantic} {Segmentation},'' \emph{arXiv:1704.06857 [cs]}, Apr. 2017,
  arXiv: 1704.06857. [Online]. Available: \url{http://arxiv.org/abs/1704.06857}
\BIBentrySTDinterwordspacing

\bibitem{zhang_how_2020}
\BIBentryALTinterwordspacing
X.~Zhang, L.~Han, L.~Han, and L.~Zhu, ``\BIBforeignlanguage{en}{How {Well} {Do}
  {Deep} {Learning}-{Based} {Methods} for {Land} {Cover} {Classification} and
  {Object} {Detection} {Perform} on {High} {Resolution} {Remote} {Sensing}
  {Imagery}?}'' \emph{\BIBforeignlanguage{en}{Remote Sensing}}, vol.~12, no.~3,
  p. 417, Jan. 2020, number: 3 Publisher: Multidisciplinary Digital Publishing
  Institute. [Online]. Available: \url{https://www.mdpi.com/2072-4292/12/3/417}
\BIBentrySTDinterwordspacing

\bibitem{ball_comprehensive_2017}
\BIBentryALTinterwordspacing
J.~E. Ball, D.~T. Anderson, and C.~S. Chan, ``A {Comprehensive} {Survey} of
  {Deep} {Learning} in {Remote} {Sensing}: {Theories}, {Tools} and {Challenges}
  for the {Community},'' \emph{Journal of Applied Remote Sensing}, vol.~11,
  no.~04, p.~1, Sep. 2017, arXiv: 1709.00308. [Online]. Available:
  \url{http://arxiv.org/abs/1709.00308}
\BIBentrySTDinterwordspacing

\bibitem{romero_unsupervised_2016}
\BIBentryALTinterwordspacing
A.~Romero, C.~Gatta, and G.~Camps-Valls, ``\BIBforeignlanguage{en}{Unsupervised
  {Deep} {Feature} {Extraction} for {Remote} {Sensing} {Image}
  {Classification}},'' \emph{\BIBforeignlanguage{en}{IEEE Transactions on
  Geoscience and Remote Sensing}}, vol.~54, no.~3, pp. 1349--1362, Mar. 2016.
  [Online]. Available: \url{http://ieeexplore.ieee.org/document/7293195/}
\BIBentrySTDinterwordspacing

\bibitem{hatano_image_2020}
T.~Hatano, T.~Tsuneda, Y.~Suzuki, K.~Shintani, and S.~Yamane, ``Image
  {Classification} with {Additional} {Non}-decision {Labels} using
  {Self}-supervised learning and {GAN},'' in \emph{2020 {Eighth}
  {International} {Symposium} on {Computing} and {Networking} {Workshops}
  ({CANDARW})}.\hskip 1em plus 0.5em minus 0.4em\relax IEEE, 2020, pp.
  125--129.

\bibitem{li_multi-task_2020}
Y.~Li, J.~Chen, and Y.~Zheng, ``A multi-task self-supervised learning framework
  for scopy images,'' in \emph{2020 {IEEE} 17th international symposium on
  biomedical imaging ({ISBI})}.\hskip 1em plus 0.5em minus 0.4em\relax IEEE,
  2020, pp. 2005--2009.

\bibitem{lan_albert_2019}
Z.~Lan, M.~Chen, S.~Goodman, K.~Gimpel, P.~Sharma, and R.~Soricut, ``Albert:
  {A} lite bert for self-supervised learning of language representations,''
  \emph{arXiv preprint arXiv:1909.11942}, 2019.

\bibitem{leiter_chatgpt_2023}
\BIBentryALTinterwordspacing
C.~Leiter, R.~Zhang, Y.~Chen, J.~Belouadi, D.~Larionov, V.~Fresen, and S.~Eger,
  ``\BIBforeignlanguage{en}{{ChatGPT}: {A} {Meta}-{Analysis} after 2.5
  {Months}},'' Feb. 2023. [Online]. Available:
  \url{https://arxiv.org/abs/2302.13795v1}
\BIBentrySTDinterwordspacing

\bibitem{misra_self-supervised_2020}
I.~Misra and L.~v.~d. Maaten, ``Self-supervised learning of pretext-invariant
  representations,'' in \emph{Proceedings of the {IEEE}/{CVF} conference on
  computer vision and pattern recognition}, 2020, pp. 6707--6717.

\bibitem{mitash_self-supervised_2017}
C.~Mitash, K.~E. Bekris, and A.~Boularias, ``A self-supervised learning system
  for object detection using physics simulation and multi-view pose
  estimation,'' in \emph{2017 {IEEE}/{RSJ} {International} {Conference} on
  {Intelligent} {Robots} and {Systems} ({IROS})}.\hskip 1em plus 0.5em minus
  0.4em\relax IEEE, 2017, pp. 545--551.

\bibitem{alosaimi_self-supervised_2023}
\BIBentryALTinterwordspacing
N.~Alosaimi, H.~Alhichri, Y.~Bazi, B.~Ben~Youssef, and N.~Alajlan,
  ``\BIBforeignlanguage{en}{Self-supervised learning for remote sensing scene
  classification under the few shot scenario},''
  \emph{\BIBforeignlanguage{en}{Scientific Reports}}, vol.~13, no.~1, p. 433,
  Jan. 2023, number: 1 Publisher: Nature Publishing Group. [Online]. Available:
  \url{https://www.nature.com/articles/s41598-022-27313-5}
\BIBentrySTDinterwordspacing

\bibitem{tao_remote_2022}
C.~Tao, J.~Qi, W.~Lu, H.~Wang, and H.~Li, ``Remote {Sensing} {Image} {Scene}
  {Classification} {With} {Self}-{Supervised} {Paradigm} {Under} {Limited}
  {Labeled} {Samples},'' \emph{IEEE Geoscience and Remote Sensing Letters},
  vol.~19, pp. 1--5, 2022, conference Name: IEEE Geoscience and Remote Sensing
  Letters.

\bibitem{zhao_when_2020}
\BIBentryALTinterwordspacing
Z.~Zhao, Z.~Luo, J.~Li, C.~Chen, and Y.~Piao, ``\BIBforeignlanguage{en}{When
  {Self}-{Supervised} {Learning} {Meets} {Scene} {Classification}: {Remote}
  {Sensing} {Scene} {Classification} {Based} on a {Multitask} {Learning}
  {Framework}},'' \emph{\BIBforeignlanguage{en}{Remote Sensing}}, vol.~12,
  no.~20, p. 3276, Jan. 2020, number: 20 Publisher: Multidisciplinary Digital
  Publishing Institute. [Online]. Available:
  \url{https://www.mdpi.com/2072-4292/12/20/3276}
\BIBentrySTDinterwordspacing

\bibitem{dong_self-supervised_2020}
\BIBentryALTinterwordspacing
H.~Dong, W.~Ma, Y.~Wu, J.~Zhang, and L.~Jiao,
  ``\BIBforeignlanguage{en}{Self-{Supervised} {Representation} {Learning} for
  {Remote} {Sensing} {Image} {Change} {Detection} {Based} on {Temporal}
  {Prediction}},'' \emph{\BIBforeignlanguage{en}{Remote Sensing}}, vol.~12,
  no.~11, p. 1868, Jan. 2020, number: 11 Publisher: Multidisciplinary Digital
  Publishing Institute. [Online]. Available:
  \url{https://www.mdpi.com/2072-4292/12/11/1868}
\BIBentrySTDinterwordspacing

\bibitem{zhang_self-supervised_2022}
X.~Zhang, L.~Han, T.~Sobeih, L.~Lappin, M.~A. Lee, A.~Howard, and A.~Kisdi,
  ``The {Self}-{Supervised} {Spectral}–{Spatial} {Vision} {Transformer}
  {Network} for {Accurate} {Prediction} of {Wheat} {Nitrogen} {Status} from
  {UAV} {Imagery},'' \emph{Remote Sensing}, vol.~14, no.~6, p. 1400, 2022,
  publisher: MDPI.

\bibitem{he_masked_2021}
\BIBentryALTinterwordspacing
K.~He, X.~Chen, S.~Xie, Y.~Li, P.~Dollár, and R.~Girshick, ``Masked
  {Autoencoders} {Are} {Scalable} {Vision} {Learners},'' \emph{arXiv:2111.06377
  [cs]}, Dec. 2021, arXiv: 2111.06377. [Online]. Available:
  \url{http://arxiv.org/abs/2111.06377}
\BIBentrySTDinterwordspacing

\bibitem{komodakis_unsupervised_2018}
N.~Komodakis and S.~Gidaris, ``Unsupervised representation learning by
  predicting image rotations,'' in \emph{International {Conference} on
  {Learning} {Representations} ({ICLR})}, 2018.

\bibitem{imani_overview_2020}
\BIBentryALTinterwordspacing
M.~Imani and H.~Ghassemian, ``\BIBforeignlanguage{en}{An overview on spectral
  and spatial information fusion for hyperspectral image classification:
  {Current} trends and challenges},'' \emph{\BIBforeignlanguage{en}{Information
  Fusion}}, vol.~59, pp. 59--83, Jul. 2020. [Online]. Available:
  \url{https://www.sciencedirect.com/science/article/pii/S1566253519307857}
\BIBentrySTDinterwordspacing

\bibitem{fauvel_spectral_2007}
M.~Fauvel, J.~Chanussot, J.~A. Benediktsson, and J.~R. Sveinsson, ``Spectral
  and spatial classification of hyperspectral data using {SVMs} and
  morphological profiles,'' in \emph{2007 {IEEE} {International} {Geoscience}
  and {Remote} {Sensing} {Symposium}}, Jul. 2007, pp. 4834--4837.

\bibitem{lee_svm-based_2015}
W.~Lee, B.~Park, and K.~Han, ``Svm-based classification of diffusion tensor
  imaging data for diagnosing alzheimer’s disease and mild cognitive
  impairment,'' in \emph{International {Conference} on {Intelligent}
  {Computing}}.\hskip 1em plus 0.5em minus 0.4em\relax Springer, 2015, pp.
  489--499.

\bibitem{belgiu_random_2016}
\BIBentryALTinterwordspacing
M.~Belgiu and L.~Drăguţ, ``\BIBforeignlanguage{en}{Random forest in remote
  sensing: {A} review of applications and future directions},''
  \emph{\BIBforeignlanguage{en}{ISPRS Journal of Photogrammetry and Remote
  Sensing}}, vol. 114, pp. 24--31, Apr. 2016. [Online]. Available:
  \url{https://linkinghub.elsevier.com/retrieve/pii/S0924271616000265}
\BIBentrySTDinterwordspacing

\bibitem{chasmer_decision-tree_2014}
L.~Chasmer, C.~Hopkinson, T.~Veness, W.~Quinton, and J.~Baltzer, ``A
  decision-tree classification for low-lying complex land cover types within
  the zone of discontinuous permafrost,'' \emph{Remote Sensing of Environment},
  vol. 143, no.~10, pp. 73--84, 2014.

\bibitem{friedl_decision_1997}
M.~A. Friedl and C.~E. Brodley, ``Decision tree classification of land cover
  from remotely sensed data,'' \emph{Remote Sensing of Environment}, vol.~61,
  no.~3, pp. 399--409, 1997.

\bibitem{ball_special_2018}
\BIBentryALTinterwordspacing
J.~E. Ball, D.~T. Anderson, and C.~S. Chan, ``\BIBforeignlanguage{en}{Special
  {Section} {Guest} {Editorial}: {Feature} and {Deep} {Learning} in {Remote}
  {Sensing} {Applications}},'' \emph{\BIBforeignlanguage{en}{Journal of Applied
  Remote Sensing}}, vol.~11, no.~04, p.~1, Jan. 2018. [Online]. Available:
  \url{https://www.spiedigitallibrary.org/journals/journal-of-applied-remote-sensing/volume-11/issue-04/042601/Special-Section-Guest-Editorial--Feature-and-Deep-Learning-in/10.1117/1.JRS.11.042601.full}
\BIBentrySTDinterwordspacing

\bibitem{brown_dynamic_2022}
\BIBentryALTinterwordspacing
C.~F. Brown, S.~P. Brumby, B.~Guzder-Williams, T.~Birch, S.~B. Hyde,
  J.~Mazzariello, W.~Czerwinski, V.~J. Pasquarella, R.~Haertel,
  S.~Ilyushchenko, K.~Schwehr, M.~Weisse, F.~Stolle, C.~Hanson, O.~Guinan,
  R.~Moore, and A.~M. Tait, ``\BIBforeignlanguage{en}{Dynamic {World}, {Near}
  real-time global 10 m land use land cover mapping},''
  \emph{\BIBforeignlanguage{en}{Scientific Data}}, vol.~9, no.~1, p. 251, Jun.
  2022, number: 1 Publisher: Nature Publishing Group. [Online]. Available:
  \url{https://www.nature.com/articles/s41597-022-01307-4}
\BIBentrySTDinterwordspacing

\bibitem{wang_self-supervised_2022}
Y.~Wang, C.~M. Albrecht, N.~A.~A. Braham, L.~Mou, and X.~X. Zhu,
  ``Self-{Supervised} {Learning} in {Remote} {Sensing}: {A} review,''
  \emph{IEEE Geoscience and Remote Sensing Magazine}, vol.~10, no.~4, pp.
  213--247, Dec. 2022, conference Name: IEEE Geoscience and Remote Sensing
  Magazine.

\bibitem{bruzzone2001unsupervised}
L.~Bruzzone and D.~F. Prieto, ``Unsupervised retraining of a maximum likelihood
  classifier for the analysis of multitemporal remote sensing images,''
  \emph{IEEE Transactions on Geoscience and Remote Sensing}, vol.~39, no.~2,
  pp. 456--460, 2001.

\bibitem{congalton1991review}
R.~G. Congalton, ``A review of assessing the accuracy of classifications of
  remotely sensed data,'' \emph{Remote sensing of environment}, vol.~37, no.~1,
  pp. 35--46, 1991.

\bibitem{ball_isodata:_1965}
G.~H. Ball and J.~Hall, ``{ISODATA}: {A} novel method for data analysis and
  pattern classification,'' 1965.

\bibitem{kanungo_efficient_2002}
\BIBentryALTinterwordspacing
T.~Kanungo, D.~Mount, N.~Netanyahu, C.~Piatko, R.~Silverman, and A.~Wu,
  ``\BIBforeignlanguage{en}{An efficient k-means clustering algorithm: analysis
  and implementation},'' \emph{\BIBforeignlanguage{en}{IEEE Transactions on
  Pattern Analysis and Machine Intelligence}}, vol.~24, no.~7, pp. 881--892,
  Jul. 2002. [Online]. Available:
  \url{http://ieeexplore.ieee.org/document/1017616/}
\BIBentrySTDinterwordspacing

\bibitem{zhang_crop_2016}
\BIBentryALTinterwordspacing
X.~Zhang, M.~Zhang, Y.~Zheng, and B.~Wu, ``Crop {Mapping} {Using} {PROBA}-{V}
  {Time} {Series} {Data} at the {Yucheng} and {Hongxing} {Farm} in {China},''
  \emph{Remote Sensing}, vol.~8, no.~11, p. 915, 2016. [Online]. Available:
  \url{http://www.mdpi.com/2072-4292/8/11/915 DE ID - 337}
\BIBentrySTDinterwordspacing

\bibitem{zhang_spectralspatial_2016}
H.~Zhang, H.~Zhai, L.~Zhang, and P.~Li, ``Spectral–spatial sparse subspace
  clustering for hyperspectral remote sensing images,'' \emph{IEEE Transactions
  on Geoscience and Remote Sensing}, vol.~54, no.~6, pp. 3672--3684, 2016,
  publisher: IEEE.

\bibitem{doersch_unsupervised_2015}
C.~Doersch, A.~Gupta, and A.~A. Efros, ``Unsupervised visual representation
  learning by context prediction,'' in \emph{Proceedings of the {IEEE}
  international conference on computer vision}, 2015, pp. 1422--1430.

\bibitem{noroozi_unsupervised_2016}
M.~Noroozi and P.~Favaro, ``Unsupervised learning of visual representations by
  solving jigsaw puzzles,'' in \emph{European conference on computer
  vision}.\hskip 1em plus 0.5em minus 0.4em\relax Springer, 2016, pp. 69--84.

\bibitem{alexey_discriminative_2016}
D.~Alexey, P.~Fischer, J.~Tobias, M.~R. Springenberg, and T.~Brox,
  ``Discriminative, unsupervised feature learning with exemplar convolutional,
  neural networks,'' \emph{IEEE TPAMI}, vol.~38, no.~9, pp. 1734--1747, 2016.

\bibitem{arora_theoretical_2019}
\BIBentryALTinterwordspacing
S.~Arora, H.~Khandeparkar, M.~Khodak, O.~Plevrakis, and N.~Saunshi, ``A
  {Theoretical} {Analysis} of {Contrastive} {Unsupervised} {Representation}
  {Learning},'' \emph{arXiv:1902.09229 [cs, stat]}, Feb. 2019, arXiv:
  1902.09229. [Online]. Available: \url{http://arxiv.org/abs/1902.09229}
\BIBentrySTDinterwordspacing

\bibitem{caron_emerging_2021}
\BIBentryALTinterwordspacing
M.~Caron, H.~Touvron, I.~Misra, H.~Jégou, J.~Mairal, P.~Bojanowski, and
  A.~Joulin, ``Emerging {Properties} in {Self}-{Supervised} {Vision}
  {Transformers},'' \emph{arXiv:2104.14294 [cs]}, May 2021, arXiv: 2104.14294.
  [Online]. Available: \url{http://arxiv.org/abs/2104.14294}
\BIBentrySTDinterwordspacing

\bibitem{grill_bootstrap_2020}
\BIBentryALTinterwordspacing
J.-B. Grill, F.~Strub, F.~Altché, C.~Tallec, P.~H. Richemond, E.~Buchatskaya,
  C.~Doersch, B.~A. Pires, Z.~D. Guo, M.~G. Azar, B.~Piot, K.~Kavukcuoglu,
  R.~Munos, and M.~Valko, ``Bootstrap your own latent: {A} new approach to
  self-supervised {Learning},'' \emph{arXiv:2006.07733 [cs, stat]}, Sep. 2020,
  arXiv: 2006.07733. [Online]. Available: \url{http://arxiv.org/abs/2006.07733}
\BIBentrySTDinterwordspacing

\bibitem{vincent_stacked_2010}
P.~Vincent, H.~Larochelle, I.~Lajoie, Y.~Bengio, P.-A. Manzagol, and L.~Bottou,
  ``Stacked denoising autoencoders: {Learning} useful representations in a deep
  network with a local denoising criterion.'' \emph{Journal of machine learning
  research}, vol.~11, no.~12, 2010.

\bibitem{goodfellow_generative_2014}
\BIBentryALTinterwordspacing
I.~Goodfellow, J.~Pouget-Abadie, M.~Mirza, B.~Xu, D.~Warde-Farley, S.~Ozair,
  A.~Courville, and Y.~Bengio, ``Generative {Adversarial} {Nets},'' in
  \emph{Advances in {Neural} {Information} {Processing} {Systems} 27},
  Z.~Ghahramani, M.~Welling, C.~Cortes, N.~D. Lawrence, and K.~Q. Weinberger,
  Eds.\hskip 1em plus 0.5em minus 0.4em\relax Curran Associates, Inc., 2014,
  pp. 2672--2680. [Online]. Available:
  \url{http://papers.nips.cc/paper/5423-generative-adversarial-nets.pdf}
\BIBentrySTDinterwordspacing

\bibitem{arjovsky_wasserstein_2017}
\BIBentryALTinterwordspacing
M.~Arjovsky, S.~Chintala, and L.~Bottou, ``Wasserstein {GAN},''
  \emph{arXiv:1701.07875 [cs, stat]}, Dec. 2017, arXiv: 1701.07875. [Online].
  Available: \url{http://arxiv.org/abs/1701.07875}
\BIBentrySTDinterwordspacing

\bibitem{chen_improved_2020}
X.~Chen, H.~Fan, R.~Girshick, and K.~He, ``Improved baselines with momentum
  contrastive learning,'' \emph{arXiv preprint arXiv:2003.04297}, 2020.

\bibitem{chen_empirical_2021}
\BIBentryALTinterwordspacing
X.~Chen, S.~Xie, and K.~He, ``An {Empirical} {Study} of {Training}
  {Self}-{Supervised} {Vision} {Transformers},'' \emph{arXiv:2104.02057 [cs]},
  Aug. 2021, arXiv: 2104.02057. [Online]. Available:
  \url{http://arxiv.org/abs/2104.02057}
\BIBentrySTDinterwordspacing

\bibitem{he_momentum_2020}
K.~He, H.~Fan, Y.~Wu, S.~Xie, and R.~Girshick, ``Momentum contrast for
  unsupervised visual representation learning,'' in \emph{Proceedings of the
  {IEEE}/{CVF} {Conference} on {Computer} {Vision} and {Pattern}
  {Recognition}}, 2020, pp. 9729--9738.

\bibitem{chen_exploring_2020}
\BIBentryALTinterwordspacing
X.~Chen and K.~He, ``Exploring {Simple} {Siamese} {Representation}
  {Learning},'' Nov. 2020, arXiv:2011.10566 [cs]. [Online]. Available:
  \url{http://arxiv.org/abs/2011.10566}
\BIBentrySTDinterwordspacing

\bibitem{wen_rotation_2021}
Z.~Wen, Z.~Liu, S.~Zhang, and Q.~Pan, ``Rotation awareness based
  self-supervised learning for {SAR} target recognition with limited training
  samples,'' \emph{IEEE Transactions on Image Processing}, vol.~30, pp.
  7266--7279, 2021, publisher: IEEE.

\bibitem{singh_self-supervised_2018}
S.~Singh, A.~Batra, G.~Pang, L.~Torresani, S.~Basu, M.~Paluri, and C.~V.
  Jawahar, ``Self-{Supervised} {Feature} {Learning} for {Semantic}
  {Segmentation} of {Overhead} {Imagery}.'' in \emph{{BMVC}}, vol.~1, 2018,
  p.~4, issue: 2.

\bibitem{geng_multi-view_2022}
W.~Geng, W.~Zhou, and S.~Jin, ``Multi-view urban scene classification with a
  complementary-information learning model,'' \emph{Photogrammetric Engineering
  \& Remote Sensing}, vol.~88, no.~1, pp. 65--72, 2022, publisher: American
  Society for Photogrammetry and Remote Sensing.

\bibitem{rao_transferable_2022}
W.~Rao, Y.~Qu, L.~Gao, X.~Sun, Y.~Wu, and B.~Zhang, ``Transferable network with
  {Siamese} architecture for anomaly detection in hyperspectral images,''
  \emph{International Journal of Applied Earth Observation and Geoinformation},
  vol. 106, p. 102669, 2022, publisher: Elsevier.

\bibitem{zhang_semisupervised_2022}
L.~Zhang, W.~Lu, J.~Zhang, and H.~Wang, ``A {Semisupervised} {Convolution}
  {Neural} {Network} for {Partial} {Unlabeled} {Remote}-{Sensing} {Image}
  {Segmentation},'' \emph{IEEE Geoscience and Remote Sensing Letters}, vol.~19,
  pp. 1--5, 2022, publisher: IEEE.

\bibitem{jean_tile2vec_2018}
\BIBentryALTinterwordspacing
N.~Jean, S.~Wang, A.~Samar, G.~Azzari, D.~Lobell, and S.~Ermon, ``{Tile2Vec}:
  {Unsupervised} representation learning for spatially distributed data,'' May
  2018, arXiv:1805.02855 [cs, stat]. [Online]. Available:
  \url{http://arxiv.org/abs/1805.02855}
\BIBentrySTDinterwordspacing

\bibitem{hou_hyperspectral_2021}
S.~Hou, H.~Shi, X.~Cao, X.~Zhang, and L.~Jiao, ``Hyperspectral imagery
  classification based on contrastive learning,'' \emph{IEEE Transactions on
  Geoscience and Remote Sensing}, vol.~60, pp. 1--13, 2021, publisher: IEEE.

\bibitem{duan_self-supervised_2022}
P.~Duan, Z.~Xie, X.~Kang, and S.~Li, ``Self-supervised learning-based oil spill
  detection of hyperspectral images,'' \emph{Science China Technological
  Sciences}, vol.~65, no.~4, pp. 793--801, 2022, publisher: Springer.

\bibitem{zhu_sc-eadnet_2022}
M.~Zhu, J.~Fan, Q.~Yang, and T.~Chen, ``{SC}-{EADNet}: {A} {Self}-{Supervised}
  {Contrastive} {Efficient} {Asymmetric} {Dilated} {Network} for
  {Hyperspectral} {Image} {Classification},'' \emph{IEEE Transactions on
  Geoscience and Remote Sensing}, vol.~60, pp. 1--17, 2022, conference Name:
  IEEE Transactions on Geoscience and Remote Sensing.

\bibitem{chen_simple_2020}
\BIBentryALTinterwordspacing
T.~Chen, S.~Kornblith, M.~Norouzi, and G.~Hinton, ``A {Simple} {Framework} for
  {Contrastive} {Learning} of {Visual} {Representations},''
  \emph{arXiv:2002.05709 [cs, stat]}, Jun. 2020, arXiv: 2002.05709. [Online].
  Available: \url{http://arxiv.org/abs/2002.05709}
\BIBentrySTDinterwordspacing

\bibitem{he_deep_2015}
\BIBentryALTinterwordspacing
K.~He, X.~Zhang, S.~Ren, and J.~Sun, ``\BIBforeignlanguage{en}{Deep {Residual}
  {Learning} for {Image} {Recognition}},''
  \emph{\BIBforeignlanguage{en}{arXiv:1512.03385 [cs]}}, Dec. 2015, arXiv:
  1512.03385. [Online]. Available: \url{http://arxiv.org/abs/1512.03385}
\BIBentrySTDinterwordspacing

\bibitem{buades_non-local_2005}
A.~Buades, B.~Coll, and J.-M. Morel, ``A non-local algorithm for image
  denoising,'' in \emph{2005 {IEEE} {Computer} {Society} {Conference} on
  {Computer} {Vision} and {Pattern} {Recognition} ({CVPR}'05)}, vol.~2.\hskip
  1em plus 0.5em minus 0.4em\relax IEEE, 2005, pp. 60--65.

\bibitem{ba_layer_2016}
\BIBentryALTinterwordspacing
J.~L. Ba, J.~R. Kiros, and G.~E. Hinton, ``Layer {Normalization},''
  \emph{arXiv:1607.06450 [cs, stat]}, Jul. 2016, arXiv: 1607.06450. [Online].
  Available: \url{http://arxiv.org/abs/1607.06450}
\BIBentrySTDinterwordspacing

\bibitem{dong_attention_2021}
\BIBentryALTinterwordspacing
Y.~Dong, J.-B. Cordonnier, and A.~Loukas, ``Attention is {Not} {All} {You}
  {Need}: {Pure} {Attention} {Loses} {Rank} {Doubly} {Exponentially} with
  {Depth},'' \emph{arXiv:2103.03404 [cs]}, Mar. 2021, arXiv: 2103.03404
  version: 1. [Online]. Available: \url{http://arxiv.org/abs/2103.03404}
\BIBentrySTDinterwordspacing

\bibitem{simonyan_very_2014}
K.~Simonyan and A.~Zisserman, ``Very deep convolutional networks for
  large-scale image recognition,'' \emph{arXiv preprint arXiv:1409.1556}, 2014.

\bibitem{dosovitskiy_image_2020}
\BIBentryALTinterwordspacing
A.~Dosovitskiy, L.~Beyer, A.~Kolesnikov, D.~Weissenborn, X.~Zhai,
  T.~Unterthiner, M.~Dehghani, M.~Minderer, G.~Heigold, S.~Gelly, J.~Uszkoreit,
  and N.~Houlsby, ``An {Image} is {Worth} 16x16 {Words}: {Transformers} for
  {Image} {Recognition} at {Scale},'' \emph{arXiv:2010.11929 [cs]}, Oct. 2020,
  arXiv: 2010.11929. [Online]. Available: \url{http://arxiv.org/abs/2010.11929}
\BIBentrySTDinterwordspacing

\bibitem{sumbul_bigearthnet-mm_2021}
G.~Sumbul, A.~De~Wall, T.~Kreuziger, F.~Marcelino, H.~Costa, P.~Benevides,
  M.~Caetano, B.~Demir, and V.~Markl, ``{BigEarthNet}-{MM}: {A}
  {Large}-{Scale}, {Multimodal}, {Multilabel} {Benchmark} {Archive} for
  {Remote} {Sensing} {Image} {Classification} and {Retrieval} [{Software} and
  {Data} {Sets}],'' \emph{IEEE Geoscience and Remote Sensing Magazine}, vol.~9,
  no.~3, pp. 174--180, 2021, publisher: IEEE.

\bibitem{sumbul_bigearthnet_2020}
G.~Sumbul, J.~Kang, T.~Kreuziger, F.~Marcelino, H.~Costa, P.~Benevides,
  M.~Caetano, and B.~Demir, ``Bigearthnet deep learning models with a new
  class-nomenclature for remote sensing image understanding,'' \emph{arXiv
  preprint arXiv:2001.06372}, 2020.

\bibitem{sumbul_deep_2020}
G.~Sumbul and B.~Demİr, ``A {Deep} {Multi}-{Attention} {Driven} {Approach} for
  {Multi}-{Label} {Remote} {Sensing} {Image} {Classification},'' \emph{IEEE
  Access}, vol.~8, pp. 95\,934--95\,946, 2020, conference Name: IEEE Access.

\bibitem{loshchilov_decoupled_2017}
\BIBentryALTinterwordspacing
I.~Loshchilov and F.~Hutter, ``\BIBforeignlanguage{en}{Decoupled {Weight}
  {Decay} {Regularization}},'' Nov. 2017. [Online]. Available:
  \url{https://arxiv.org/abs/1711.05101v3}
\BIBentrySTDinterwordspacing

\bibitem{noauthor_ai4eo_nodate}
\BIBentryALTinterwordspacing
``{AI4EO}.'' [Online]. Available:
  \url{https://platform.ai4eo.eu/seeing-beyond-the-visible}
\BIBentrySTDinterwordspacing

\bibitem{prokhorenkova_catboost_2019}
\BIBentryALTinterwordspacing
L.~Prokhorenkova, G.~Gusev, A.~Vorobev, A.~V. Dorogush, and A.~Gulin,
  ``{CatBoost}: unbiased boosting with categorical features,''
  \emph{arXiv:1706.09516 [cs]}, Jan. 2019, arXiv: 1706.09516. [Online].
  Available: \url{http://arxiv.org/abs/1706.09516}
\BIBentrySTDinterwordspacing

\bibitem{goyal_accurate_2017}
P.~Goyal, P.~Dollár, R.~Girshick, P.~Noordhuis, L.~Wesolowski, A.~Kyrola,
  A.~Tulloch, Y.~Jia, and K.~He, ``Accurate, large minibatch sgd: {Training}
  imagenet in 1 hour,'' \emph{arXiv preprint arXiv:1706.02677}, 2017.

\bibitem{wightman_resnet_2021}
\BIBentryALTinterwordspacing
R.~Wightman, H.~Touvron, and H.~Jégou, ``{ResNet} strikes back: {An} improved
  training procedure in timm,'' \emph{arXiv:2110.00476 [cs]}, Oct. 2021, arXiv:
  2110.00476. [Online]. Available: \url{http://arxiv.org/abs/2110.00476}
\BIBentrySTDinterwordspacing

\bibitem{devlin_bert_2019}
\BIBentryALTinterwordspacing
J.~Devlin, M.-W. Chang, K.~Lee, and K.~Toutanova, ``{BERT}: {Pre}-training of
  {Deep} {Bidirectional} {Transformers} for {Language} {Understanding},'' May
  2019, arXiv:1810.04805 [cs]. [Online]. Available:
  \url{http://arxiv.org/abs/1810.04805}
\BIBentrySTDinterwordspacing

\bibitem{brown_language_2020}
\BIBentryALTinterwordspacing
T.~B. Brown, B.~Mann, N.~Ryder, M.~Subbiah, J.~Kaplan, P.~Dhariwal,
  A.~Neelakantan, P.~Shyam, G.~Sastry, A.~Askell, S.~Agarwal, A.~Herbert-Voss,
  G.~Krueger, T.~Henighan, R.~Child, A.~Ramesh, D.~M. Ziegler, J.~Wu,
  C.~Winter, C.~Hesse, M.~Chen, E.~Sigler, M.~Litwin, S.~Gray, B.~Chess,
  J.~Clark, C.~Berner, S.~McCandlish, A.~Radford, I.~Sutskever, and D.~Amodei,
  ``Language {Models} are {Few}-{Shot} {Learners},'' \emph{arXiv:2005.14165
  [cs]}, Jul. 2020, arXiv: 2005.14165. [Online]. Available:
  \url{http://arxiv.org/abs/2005.14165}
\BIBentrySTDinterwordspacing

\bibitem{thoppilan_lamda_2022}
\BIBentryALTinterwordspacing
R.~Thoppilan, D.~De~Freitas, J.~Hall, N.~Shazeer, A.~Kulshreshtha, H.-T. Cheng,
  A.~Jin, T.~Bos, L.~Baker, Y.~Du, Y.~Li, H.~Lee, H.~S. Zheng, A.~Ghafouri,
  M.~Menegali, Y.~Huang, M.~Krikun, D.~Lepikhin, J.~Qin, D.~Chen, Y.~Xu,
  Z.~Chen, A.~Roberts, M.~Bosma, V.~Zhao, Y.~Zhou, C.-C. Chang, I.~Krivokon,
  W.~Rusch, M.~Pickett, P.~Srinivasan, L.~Man, K.~Meier-Hellstern, M.~R.
  Morris, T.~Doshi, R.~D. Santos, T.~Duke, J.~Soraker, B.~Zevenbergen,
  V.~Prabhakaran, M.~Diaz, B.~Hutchinson, K.~Olson, A.~Molina, E.~Hoffman-John,
  J.~Lee, L.~Aroyo, R.~Rajakumar, A.~Butryna, M.~Lamm, V.~Kuzmina, J.~Fenton,
  A.~Cohen, R.~Bernstein, R.~Kurzweil, B.~Aguera-Arcas, C.~Cui, M.~Croak,
  E.~Chi, and Q.~Le, ``{LaMDA}: {Language} {Models} for {Dialog}
  {Applications},'' Feb. 2022, arXiv:2201.08239 [cs]. [Online]. Available:
  \url{http://arxiv.org/abs/2201.08239}
\BIBentrySTDinterwordspacing

\bibitem{diao_active_2023}
\BIBentryALTinterwordspacing
S.~Diao, P.~Wang, Y.~Lin, and T.~Zhang, ``Active {Prompting} with
  {Chain}-of-{Thought} for {Large} {Language} {Models},'' Feb. 2023,
  arXiv:2302.12246 [cs]. [Online]. Available:
  \url{http://arxiv.org/abs/2302.12246}
\BIBentrySTDinterwordspacing

\bibitem{liu_what_2021}
\BIBentryALTinterwordspacing
J.~Liu, D.~Shen, Y.~Zhang, B.~Dolan, L.~Carin, and W.~Chen, ``What {Makes}
  {Good} {In}-{Context} {Examples} for {GPT}-\$3\$?'' Jan. 2021,
  arXiv:2101.06804 [cs]. [Online]. Available:
  \url{http://arxiv.org/abs/2101.06804}
\BIBentrySTDinterwordspacing

\bibitem{saravia_prompt_2022}
\BIBentryALTinterwordspacing
E.~Saravia, ``Prompt {Engineering} {Guide},'' Dec. 2022, publication Title:
  https://github.com/dair-ai/Prompt-Engineering-Guide original-date:
  2022-12-16T16:04:50Z. [Online]. Available:
  \url{https://github.com/dair-ai/Prompt-Engineering-Guide}
\BIBentrySTDinterwordspacing

\end{thebibliography}
%

%








\end{document}